\documentclass[conference]{IEEEtran}
\IEEEoverridecommandlockouts

\usepackage{cite}
\usepackage{amsmath,amssymb,amsfonts}
\usepackage{algorithmic}
\usepackage{graphicx}
\usepackage{textcomp}
\usepackage[colorlinks]{hyperref}
\usepackage{graphicx} 
\usepackage{subcaption} 
\usepackage{booktabs}
\usepackage{tabularx}
\usepackage{xcolor}
\usepackage{float}
\def\BibTeX{{\rm B\kern-.05em{\sc i\kern-.025em b}\kern-.08em
    T\kern-.1667em\lower.7ex\hbox{E}\kern-.125emX}}

\begin{document}

\title{On-device Large Multi-modal Agent for \\ Human Activity Recognition\\


}

\author{\IEEEauthorblockN{Md Shakhrul Iman Siam}
\IEEEauthorblockA{
\textit{The Ohio State University}\\
Columbus, Ohio, USA \\
siam.5@osu.edu}
\and
\IEEEauthorblockN{Ishtiaque Ahmed Showmik}
\IEEEauthorblockA{
\textit{The Ohio State University}\\
Columbus, Ohio, USA \\
showmik.1@osu.edu}
\and
\IEEEauthorblockN{Guanqun Song}
\IEEEauthorblockA{
\textit{The Ohio State University}\\
Columbus, Ohio, USA \\
song.2107@osu.edu}
\and
\IEEEauthorblockN{Ting Zhu}
\IEEEauthorblockA{
\textit{The Ohio State University}\\
Columbus, Ohio, USA \\
zhu.3445@osu.edu}
}

\maketitle

\begin{abstract}
Human Activity Recognition (HAR) has been an active area of research, with applications ranging from healthcare to smart environments. The recent advancements in Large Language Models (LLMs) have opened new possibilities to leverage their capabilities in HAR, enabling not just activity classification but also interpretability and human-like interaction. In this paper, we present a Large Multi-Modal Agent designed for HAR, which integrates the power of LLMs to enhance both performance and user engagement.
The proposed framework not only delivers activity classification but also bridges the gap between technical outputs and user-friendly insights through its reasoning and question-answering capabilities. 
We conduct extensive evaluations using widely adopted HAR datasets, including HHAR, Shoaib, Motionsense to assess the performance of out framework. The results demonstrate that our model achieves high classification accuracy comparable to state-of-the-art methods while significantly improving interpretability through its reasoning and Q\&A capabilities. 
\end{abstract}

\begin{IEEEkeywords}
Human Activity Recognition, Large Language Models
\end{IEEEkeywords}

\section{Introduction}

Human Activity Recognition (HAR) has emerged as a crucial task in various applications, ranging from healthcare and fitness monitoring to smart home automation and industrial safety. Traditional approaches to HAR \cite{Attal2015Dec}, such as Random Forest (RF), Long Short-Term Memory (LSTM), and Recurrent Neural Networks (RNN), have gained popularity due to their ability to model temporal and spatial patterns in sensor data effectively\cite{informatics5020027}. However, these methods often come with limitations, including being task-specific and facing significant challenges in scalability when applied to diverse or varying sensor environments\cite{Willetts2018May,9894326}. Addressing these challenges is crucial for advancing HAR systems toward more robust and generalized performance.

Recent advancements in Large Language Models (LLMs) have revolutionized fields like computer vision and natural language processing (NLP), demonstrating their generalization and adaptability \cite{li2024personal}.
Their inherent ability to extract meaningful patterns and reason contextually positions them as powerful tools for complex problem-solving.
Despite their transformative impact in these domains, the potential of LLMs remains relatively underexplored in the context of sensor data, presenting a research direction to explore in order to overcome the limitations of traditional HAR methodologies.
The flexibility and adaptability of LLMs present an opportunity to address these challenges by integrating their reasoning and interpretive capabilities with sensor-based systems.

However, integrating Large Language Models with time-series data like IMU reading is challenging because time-series data lacks the inherent semantic structure and context present in natural language or visual data, which LLMs are primarily designed to process\cite{jin2023time}. Unlike text or images, time-series data consists of numerical sequences representing changes over time, making it difficult for LLMs to interpret patterns without additional pre-processing or contextual alignment. Moreover, the high dimensionality, varying lengths, and multi-channel nature of time-series data can overwhelm standard LLM architectures, which are optimized for fixed-length tokenized inputs.

To address the gap between sensory data and language models, several research efforts \cite{Imran2024Jun,wei2025,ji2024hargpt,Li2024Oct} have attempted to integrate multimodal data into AI systems. However, these approaches face significant limitations. One of the primary challenges is aligning two fundamentally different modalities—numerical time-series data from sensors and text-based data designed for language models. While some studies have proposed methods to bridge this gap, their solutions often fall short in terms of effectiveness.
Moreover, many of these approaches are not suitable for mobile or edge devices, as they rely on server-based architectures that demand high computational resources. This dependency not only limits their real-time applicability but also raises privacy concerns, as sensitive user data must be transmitted and processed on external servers. These issues underscore the need for novel, lightweight, and privacy-preserving solutions that can effectively align sensory and textual data for robust, real-world applications.
In this paper, we propose a framework leveraging the capabilities of LLMs to classify human activities from raw sensor data while providing reasoning for its decisions. The framework is designed to support both open-ended question-and-answer tasks and close-ended classification tasks. Furthermore, we generate instruction pairs to fine-tune a smaller, mobile-friendly LLM for real-time, on-device activity recognition and reasoning, paving the way for scalable and accessible HAR solutions. 

\begin{figure}[h]
    \centering
    \includegraphics[width=0.8\linewidth]{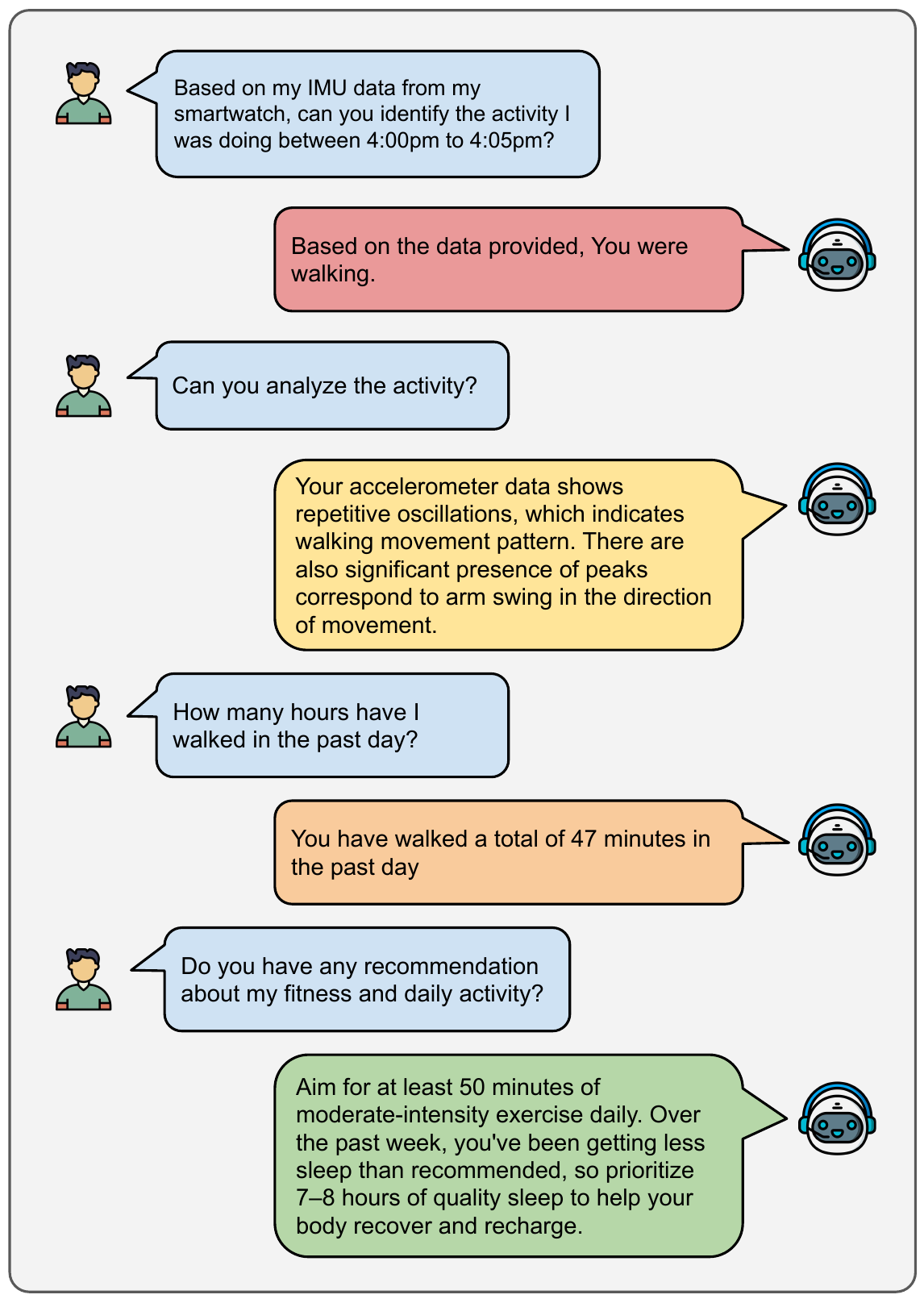} 
    \caption{The proposed framework is capable of classifying Human activity, providing reasoning, and performing QnA tasks.}
    \label{fig:res}
\end{figure}


To summarize, our work makes the following core contributions:

\begin{itemize}
    \item We developed a framework that can classify Human Activity as well as provide reasoning about its decision-making.
    \item Our framework has the question-answering capabilities.
    \item We fine-tune a computation-friendly smaller language model that can run on a mobile device. 
\end{itemize}

\section{Related Work}

\subsection{Multimodal Sensing for Smart Health}
Human Activity Recognition (HAR) models typically follow the Activity Recognition Chain \cite{bulling2014tutorial}. While initial research relied on tree-based methods \cite{yang2015deep,hammerla2016deep} and wearable inertial sensors, the scope of sensing has expanded significantly to encompass diverse, non-intrusive modalities for smart health applications.

Passive and device-free sensing approaches have been extensively explored to capture physiological and behavioral data without requiring active user intervention. For instance, Doppler radar has been utilized to recognize gestures \cite{MILLER2020100089} and monitor medication interactions \cite{MILLER2022100245} while preserving user privacy. Similarly, acoustic-based systems have been developed for passive fetal heart monitoring \cite{YAO2020100087}, and other specialized sensing frameworks have been applied to blood glucose control \cite{GAO201718} and pulmonary nodule detection \cite{8556807}.


More recently, multimodal systems fusing magnetic sensing with WiFi infrastructure have been proposed to achieve precise biometric tracking and metal detection \cite{10.1145/3769102.3770620}. Additionally, advanced visual processing techniques, such as efficient semantic segmentation on edge devices \cite{safavi2023efficientsemanticsegmentationedge}, further demonstrate the efficacy of combining diverse signals for robust activity analysis.

\subsection{Efficient IoT Infrastructure for HAR}
Deploying HAR models in real-world scenarios requires a robust and energy-efficient Internet of Things (IoT) infrastructure to ensure reliable data transmission from edge devices. Significant research has focused on optimizing connectivity in heterogeneous network environments.

To enable low-latency data transfer between incompatible wireless standards, cross-technology communication (CTC) techniques have been developed. Systems enabling concurrent high-throughput communication between WiFi and ZigBee \cite{10.1145/3210240.3210346, 10.1145/3356250.3360046} allow for real-time sensing data coordination. Furthermore, spectral efficiency improvement methods \cite{8486349, 8694952} and bi-directional communication frameworks \cite{9340574} ensure that activity data is delivered reliably even in crowded spectrums.
Addressing energy constraints, ultra-low-power methods like backscatter communication \cite{10.1145/3274783.3274846, 10.1145/3387514.3405861} and OFDM-based protocols \cite{10189210} have gained prominence. Beyond terrestrial networks, research has extended to extreme edge environments, optimizing energy efficiency for LoRaWAN in LEO satellites \cite{shergill2024energyefficientlorawanleo} and managing thermal constraints in space-based computing \cite{yuan2024heatsatellitesmeatgpus}, or even exploring global quantum communication networks \cite{gao2024optimizingglobalquantumcommunication}.

To support the massive data processing required by these systems, heterogeneous computing frameworks \cite{khatri2022heterogeneouscomputingsystems} and multiprocessing strategies for data classification and map-reduce tasks \cite{dixit2023dataclassificationmultiprocessing, qiu2023mapreducemultiprocessinglargedata} have been introduced. Moreover, recent studies emphasize the environmental impact of IoT deployment, investigating carbon neutrality \cite{yu2024achievingcarbonneutralityio} and the economic implications of hardware obsolescence \cite{cheng2024technologicalprogressobsolescenceanalyzing, gould2024environmentaleconomicimpactio} to promote sustainable smart health ecosystems.

\subsection{Security and Trustworthiness in Sensing}
As HAR systems are increasingly integrated into critical applications, ensuring security across the entire stack—from the physical layer to the operating system—is paramount.

Vulnerabilities in the physical layer have been identified across various modalities. Risks in optical communication have been revealed alongside potential defenses \cite{wire2}, and machine learning-based frameworks have been proposed to secure communication in adversarial contexts \cite{wire3, song2022mlbasedsecurelowpowercommunication}. Additionally, threats such as wireless jamming \cite{9444204, 10.1145/3395351.3399367} and invisible light attacks \cite{10.1145/3460120.3484766} highlight the need for robust defenses.

Beyond physical attacks, system-level integrity is critical. Research has analyzed security risks in OS-level virtualization \cite{ketha2025analysissecurityoslevelvirtualization} and the inner workings of operating system security \cite{kulshrestha2023innerworkingswindowssecurity}. To ensure data reliability, secure virtual file system implementations \cite{sun2023designimplementationconsiderationsvirtual} are essential. Furthermore, network-level privacy challenges in microservices architectures \cite{gopal2022securityprivacychallengesmicroservices}, blockchain scalability \cite{li2022minisculesurveyblockchainscalability}, and location tracking in 5G networks \cite{ali2023security5gnetworks} must be addressed to protect user identity while maintaining system performance \cite{285483, wire1, 9120764, 10125074}.



\subsection{Large Language Models for HAR}
Advancements in Large Language Models (LLMs) have introduced a transformative paradigm for interpreting sensor data \cite{AIOT_survey}. Zero-shot capabilities have been investigated for classifying activities directly from raw IMU data \cite{ji2024hargpt}. Other approaches leverage evolutionary strategies for unsupervised learning \cite{gao2024unsupervised} or utilize multimodal agents to interpret human activity queries \cite{Imran2024Jun}. To handle the discrepancy between numerical sensor data and text, tokenization strategies have also been proposed \cite{Li2024Oct}. As these models grow in complexity, comprehensive benchmarking remains essential for evaluating their effectiveness in edge environments \cite{ning2021benchmarkingmachinelearningfast}.

\section{Design}
\subsection{Dataset}
To facilitate our experiments, we utilized the following datasets:

\begin{figure*}[h]
\centering
    \includegraphics[width=\textwidth]{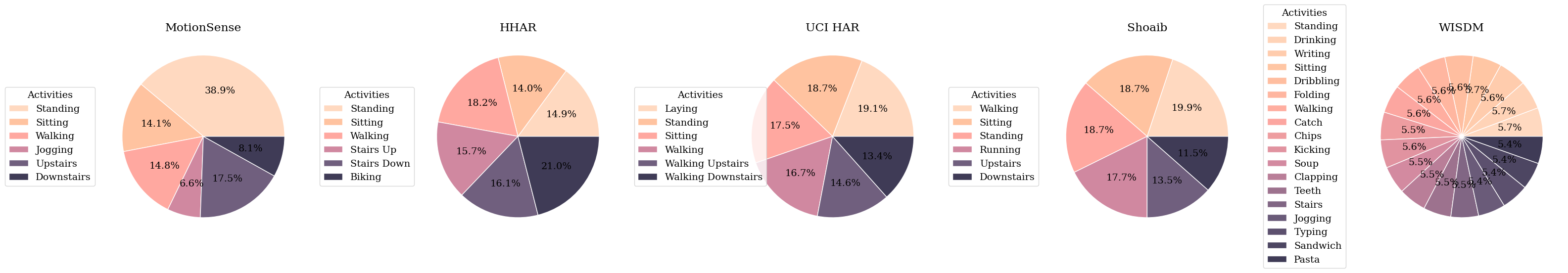} 
    \caption{Human activity labels in different datasets.}
    \label{fig:dataset-pie}
\end{figure*}

\subsubsection{Shoaib Dataset}
The Shoaib Dataset \cite{shoaib2014fusion} is for human activity recognition and sensor positioning studies. Data were collected from four Samsung Galaxy S2 smartphones placed on the arm, wrist, belt and pocket of the participants. Four male volunteers aged 25--30 performed six activities: walking, running, sitting, standing, walking upstairs, and walking downstairs. Each activity was recorded for 3--5 minutes using accelerometer, gyroscope and magnetometer sensors, capturing 3 axial readings for each modality. It is a time-series dataset comprising features as time stamp, acceleration, angular velocity, and magnetic field.

\subsubsection{Heterogeneity Dataset for Human Activity Recognition (HHAR)}
The Heterogeneity Dataset for Human Activity Recognition (HHAR) \cite{stisen2015smart} addresses sensor heterogeneities by incorporating data collected from nine participants performing six activities: biking, sitting, standing, walking, walking upstairs, and walking downstairs. Sensor readings were recorded using multiple smartphones, including Samsung Galaxy S3, Samsung Galaxy S3 Mini, and LG Nexus 4, as well as smartwatches like LG Watch and Samsung Galaxy Gear. 

\subsubsection{UCI Human Activity Recognition (HAR) Using Smartphones Dataset}
The UCI Human Activity Recognition (HAR) Using Smartphones Dataset \cite{anguita2013public}
simplifies this activity recognition through the provision of sensor data on 30 participants of ages between 19-48 years. Data preprocessing was done through the accelerometer and gyroscope of a smartphone, Samsung Galaxy S II, which was placed on the participant's waist. These participants were involved in the following activities that were to be identified: walking, walking up and down staircases, sitting, standing, laying. These data were sampled at a rate of 50 Hz, pre-processed with noise filters, and segmented into fixed-size windows of 2.56 seconds. It includes raw time-series data from each sensor as well as a processed feature set of 561 variables derived from time and frequency domains to enable comprehensive activity classification.

\subsubsection{MotionSense Dataset}
The MotionSense Dataset \cite{malekzadeh2019mobile}
features time-series for activity recognition. This was gathered using an iPhone 6s kept in the front pocket from 24 different participants representing varied demographics. Subsequent participants, performers, did all six activities: walking, jogging, sitting, standing, walking upstairs, walking downstairs, data captured from accelerometer and gyroscope sensors at a constant rate of 15 trials per participant. Sensor attributes contain a roll, pitch, user acceleration, and rotational rates-when properly segmented-represent meaningful data for activity and attribute recognition. The dataset combines both long trials of 2--3 minutes and shorter trials of 30 seconds to 1 minute, offering variability in recording conditions. 

\subsubsection{WISDM Dataset}
The WISDM Dataset \cite{wisdm_smartphone} provides activity recognition data collected from 51 participants who performed 18 activities, such as walking, jogging, sitting, standing, and climbing stairs, for three minutes each.
The dataset was captured using smartphones and smartwatches, including Nexus 5, Nexus 5X, Galaxy S6, and LG G Watch, at a sampling rate of 20 Hz. It includes raw accelerometer and gyroscope data with attributes such as subject ID, activity label, and triaxial sensor readings. Also, processed files include statistical features such as mean, standard deviation, and frequency-domain metrics, supporting diverse research in activity recognition. 

\begin{table}[h]
\renewcommand{\arraystretch}{2}
\resizebox{\columnwidth}{!}{%
\begin{tabular}{|l|l|l|l|l|}
\hline
\textbf{Dataset} & \textbf{Users} & \textbf{Class} & \textbf{Device} & \textbf{Device Placement} \\ \hline
HHAR & 9 & 6 & Smartphone, Smartwatch & Waist and Arm \\ \hline
MotionSense & 24 & 6 & Smartphone & Front Pocket \\ \hline
Shoaib & 10 & 7 & Smartphone & Arm, waist, pocket \\ \hline
UCI HAR & 30 & 6 & Smartphone & Waist \\ \hline
WISDM & 51 & 18 & Smartphones, Smartwatch & Trouser Pocket \\ \hline
\end{tabular}%
}
\caption{Summary of the Datasets}
\label{tab:dataset_details}
\end{table}

Table \ref{tab:dataset_details} shows the summary of the used datasets. 
Pie charts of Figure \ref{fig:dataset-pie} indicate the activity distribution across five datasets : MotionSense, HHAR, UCI HAR, Shoaib and WISDM. MotionSense is dominated by static activities where \textit{Standing} alone contributes 38.9\% to the data while \textit{Upstairs} and \textit{Downstairs} are underrepresented. HHAR has a more even distribution among activities, with \textit{Biking} the most dominant at 21.0\%. The UCI HAR also reflects a fairly balanced representation, \textit{Laying} and \textit{Standing} being the largest constituents with 19.1\% and 18.7\%, respectively. The Shoaib dataset has \textit{Walking} as the most dominant activity at 19.9\%, and there is a good mix of dynamic and static behaviors. Finally, WISDM features 18 activities, and all are approximately equally distributed. This ensures that multi-class classification will be diverse. 

\subsection{Data Analysis}
\subsubsection*{Time-Series Signals}
\begin{figure}[]
    \centering
    \begin{subfigure}[b]{\linewidth}
        \centering
        \includegraphics[width=\textwidth]{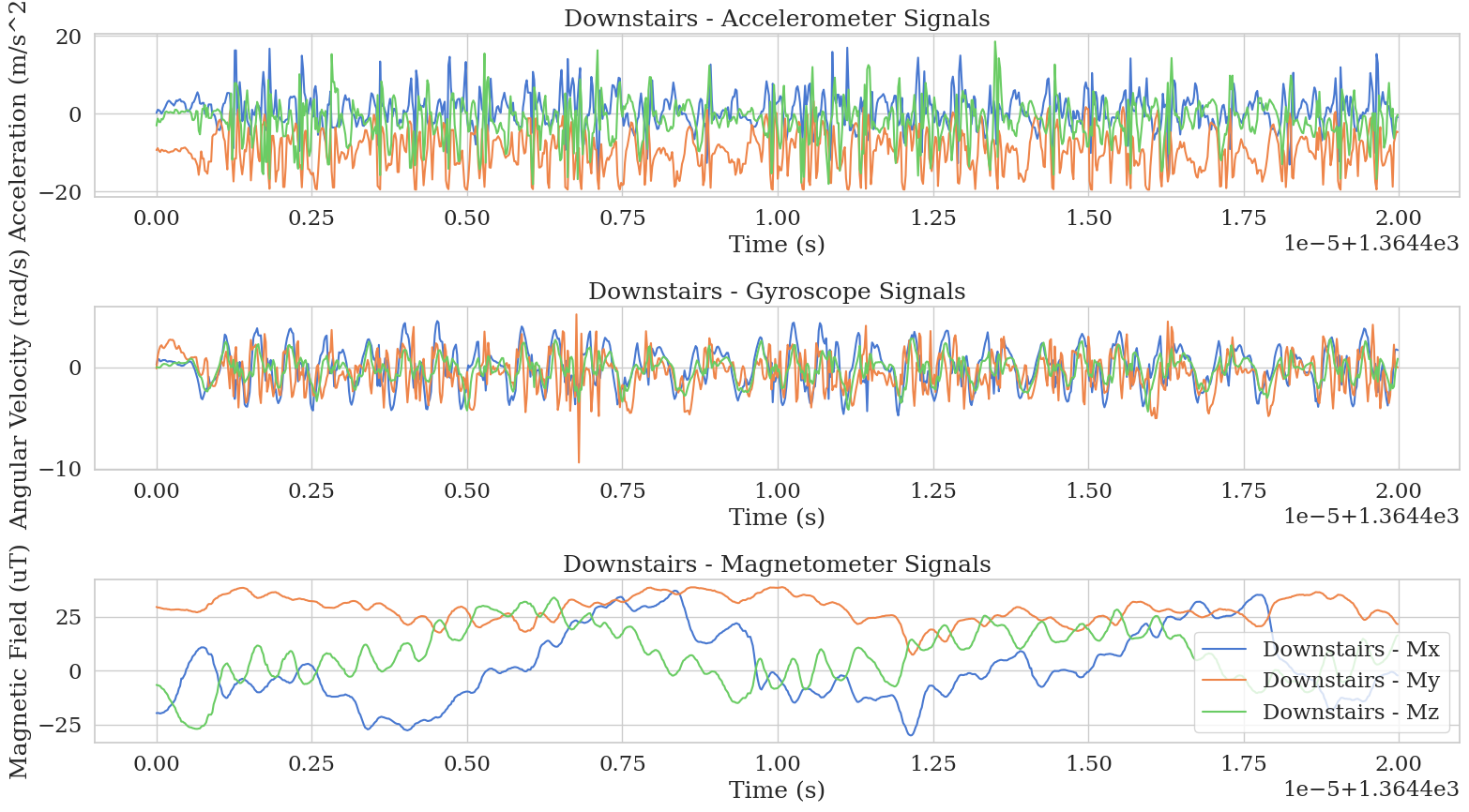} 
        \caption{Walking}
        \label{fig:walk}
    \end{subfigure}
    \hfill
    \begin{subfigure}[b]{\linewidth}
        \centering
        \includegraphics[width=\textwidth]{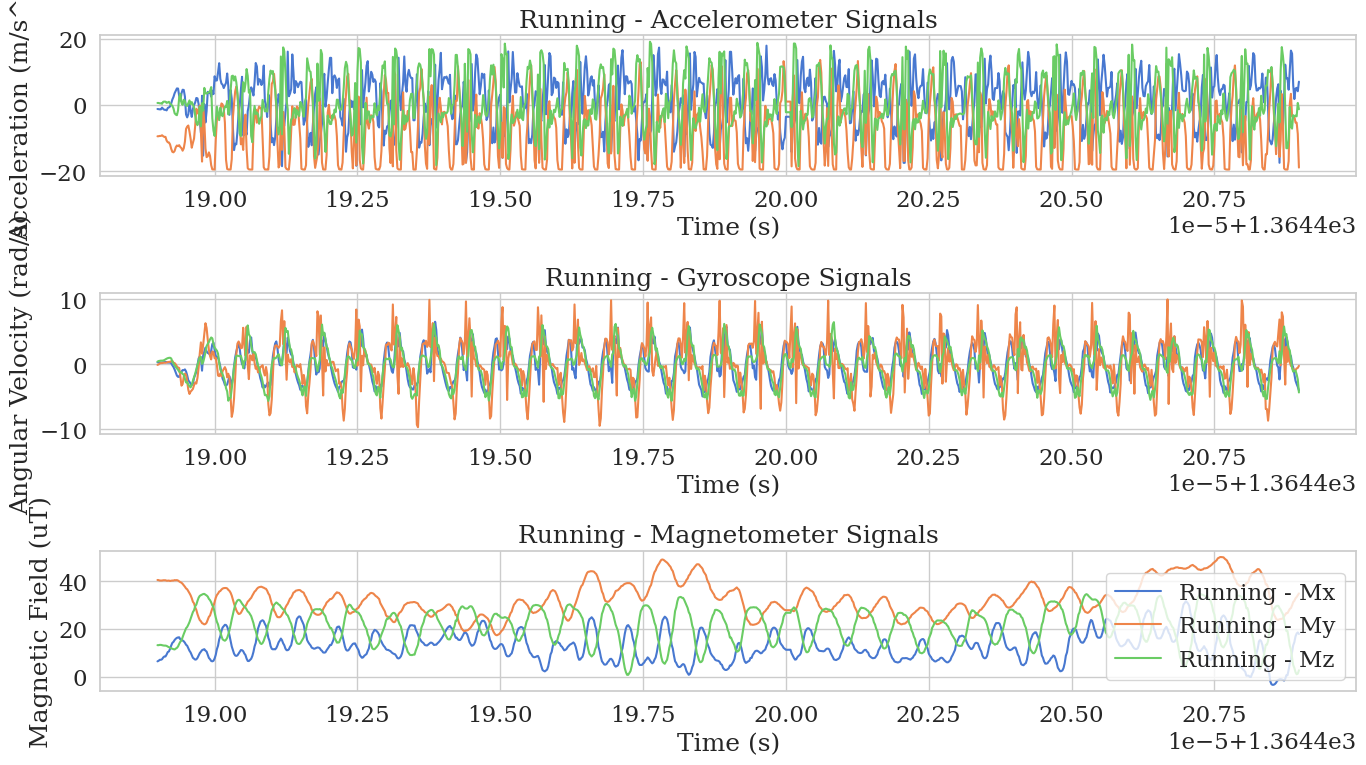} 
        \caption{Running}
        \label{fig:run}
    \end{subfigure}

    \caption{Sensor data for Walking and Running activity}
    \label{fig:act}
\end{figure}

Sensor data from accelerometers, gyroscopes, and magnetometers are presented in Fig. \ref{fig:act}. In the case of walking, accelerometer signals are periodic, highlighting the rhythmicity of the activity, while gyroscope signals reflect changes in angular velocity corresponding to limb motion. The magnetometer signals are noisier, reflecting sensitivity to environmental magnetic fields. Similarly, when running, accelerometer signals generate higher amplitude variations because of increased intensity, while the gyroscope signals will have more pronounced changes in angular velocity. These patterns underline the peculiarities of each activity in time.

\subsubsection*{Correlation Analysis}
\begin{figure}[]
    \centering
    \includegraphics[width=0.45\textwidth]{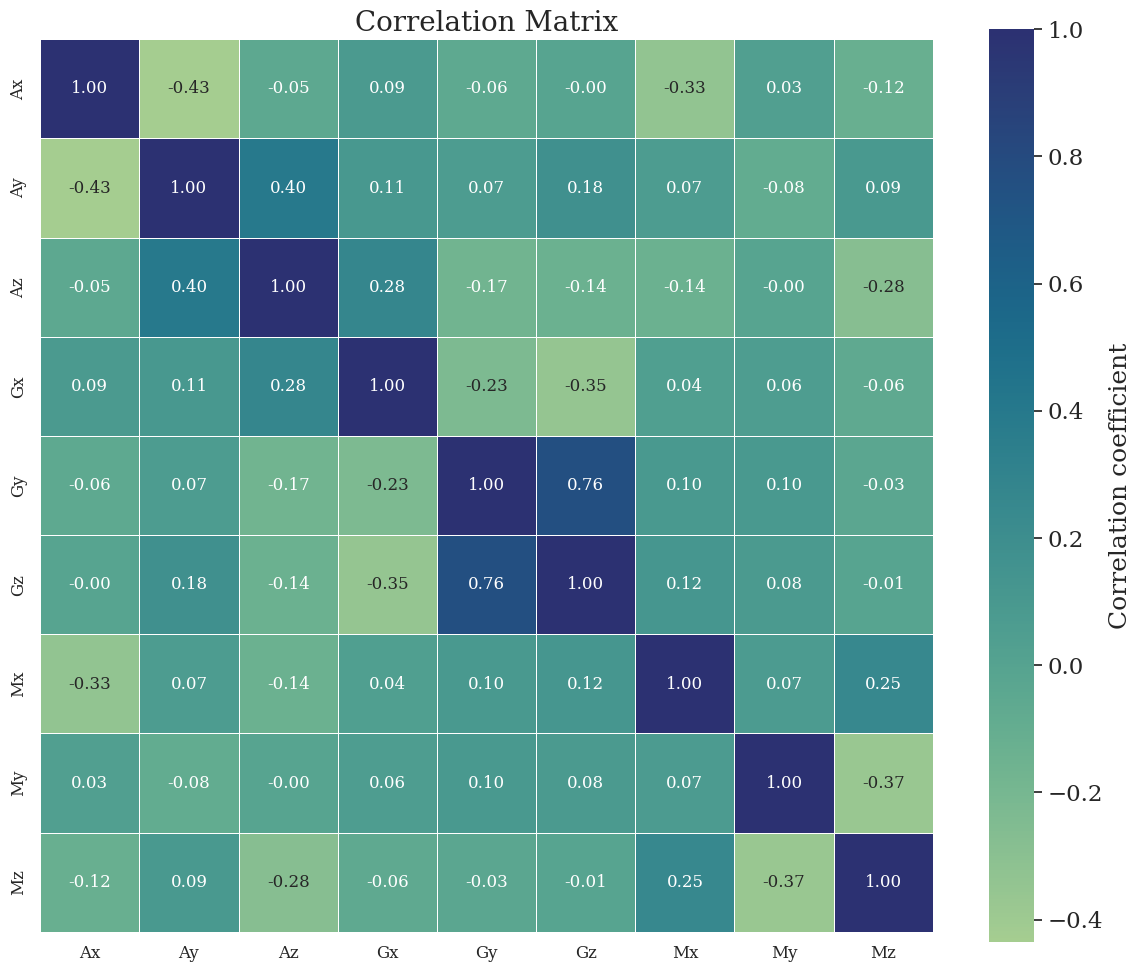} 
    \caption{Correlation Analysis.}
    \label{fig:correlation}
\end{figure}

The correlation matrix shown in Fig. \ref{fig:correlation} gives the relations among the features \(A_x, A_y, A_z, G_x, G_y, G_z, M_x, M_y, M_z\). The high values of the coefficients between gyroscope axes-for example, \( G_y \) and \( G_z \) have a correlation coefficient of 0.76-indicate coupled angular motion. The low correlations across sensor types-for example, between accelerometer and gyroscope features-indicate that the signals are complementary and will result in higher performance by making use of diverse signal characteristics.

\subsubsection*{PDF Distributions of Sensor Features}
\begin{figure}[]
    \centering
    \includegraphics[width=\linewidth]{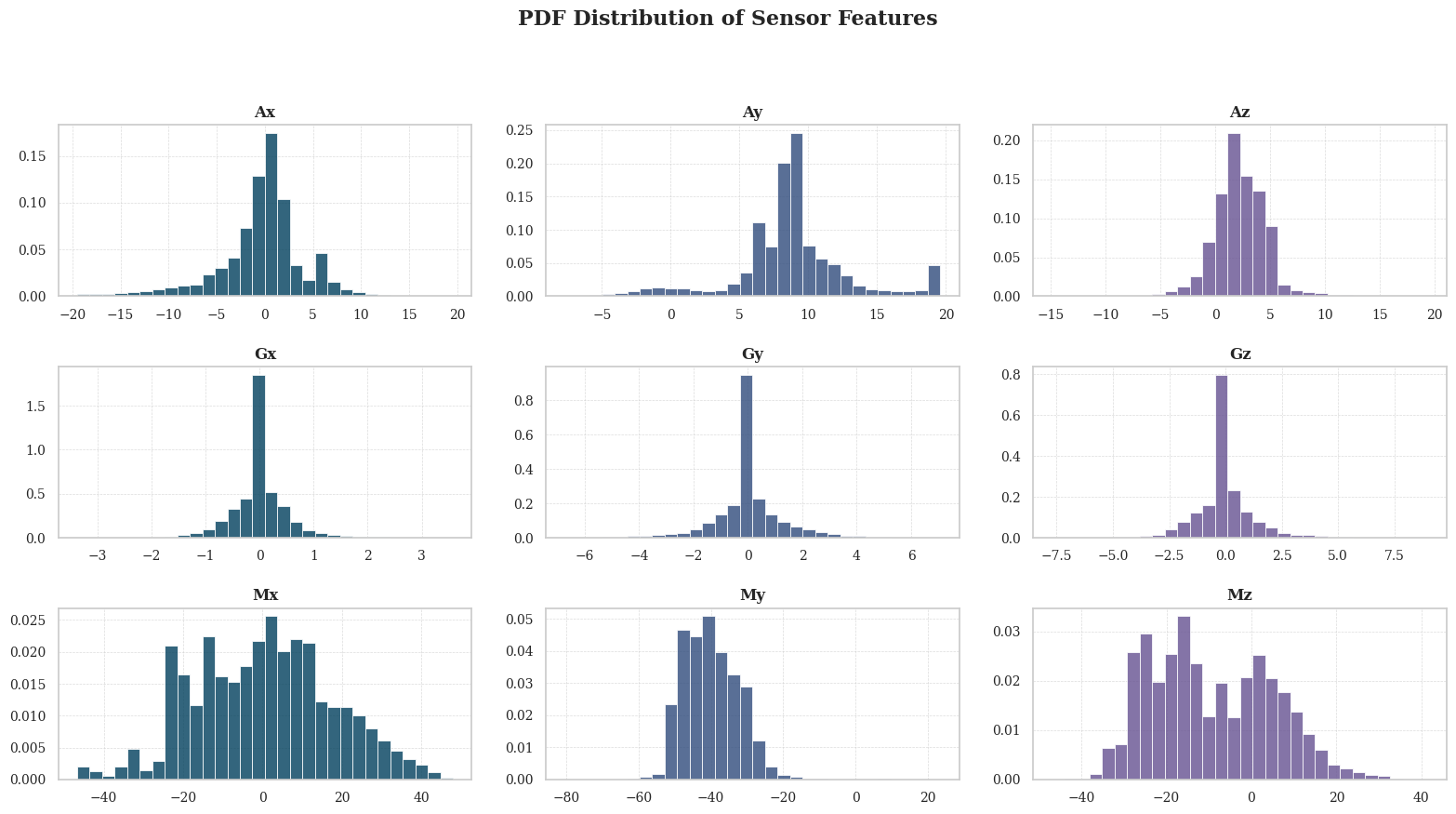} 
    \caption{PDF Distributions of Sensor Features.}
    \label{fig:pdf_distributions}
\end{figure}

The probability density functions of sensor features in Fig. \ref{fig:pdf_distributions} outline the statistical characteristics of the data. Accelerometer features such as \(A_x, A_y\), and \(A_z\) are Gaussian-like distributions centered near zero, indicating stationary movement trends. Gyroscope features such as \(G_x\), \(G_y\), and \(G_z\) are tightly clustered around zero, reflecting limited angular motion during activities. Magnetometer features such as \(M_x, M_y\), and \(M_z\) have more dispersed distributions, probably due to environmental variability.

\subsubsection*{Principal Component Analysis (PCA)}
\begin{figure}[]
    \centering
    \begin{subfigure}[b]{0.24\textwidth}
        \centering
        \includegraphics[width=\textwidth]{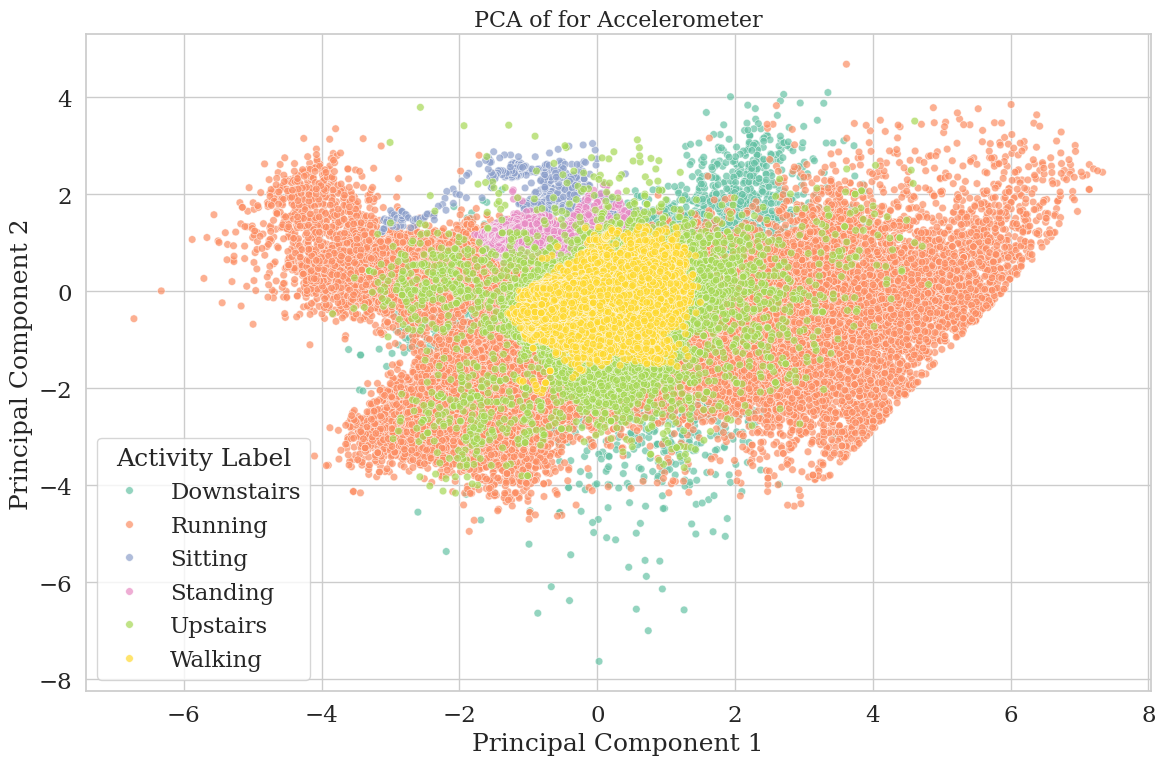} 
        \caption{PCA for Accelerometer}
        \label{pca:sub1}
    \end{subfigure}
    \hfill
    \begin{subfigure}[b]{0.24\textwidth}
        \centering
        \includegraphics[width=\textwidth]{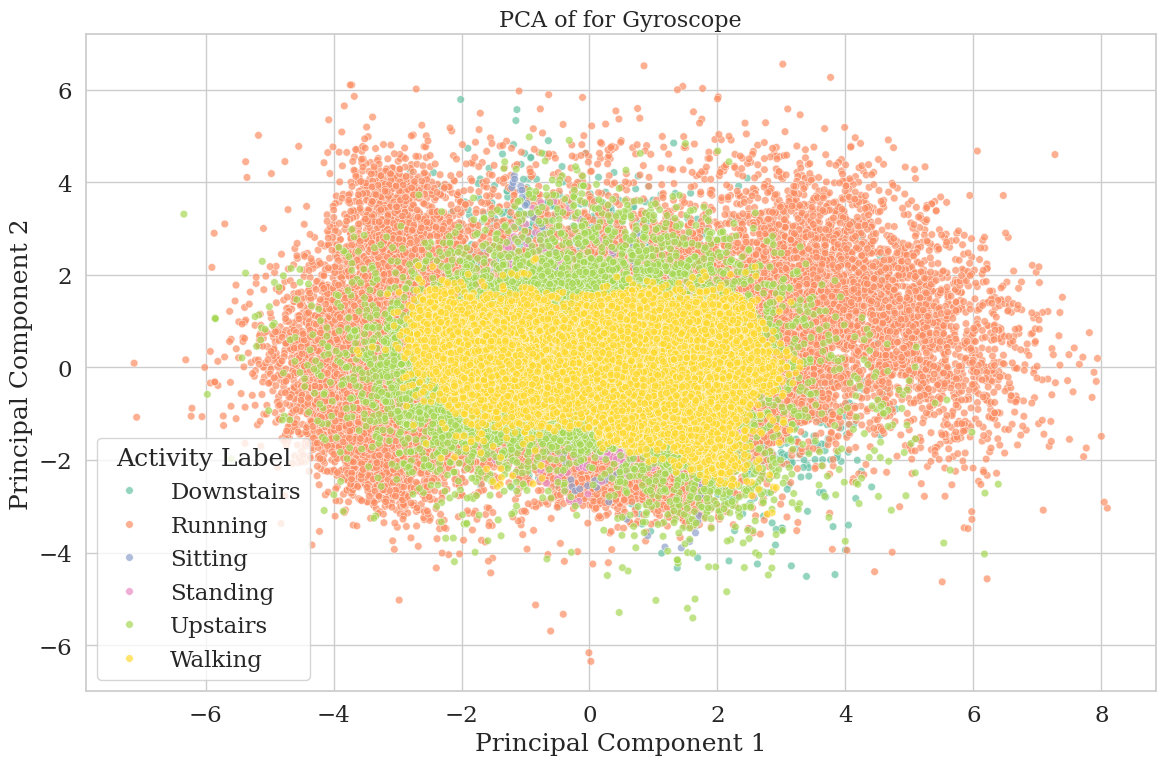} 
        \caption{PCA for Gyroscope}
        \label{pca:sub2}
        
    \end{subfigure}
    \begin{subfigure}[b]{0.25\textwidth}
        \centering
        \includegraphics[width=\textwidth]{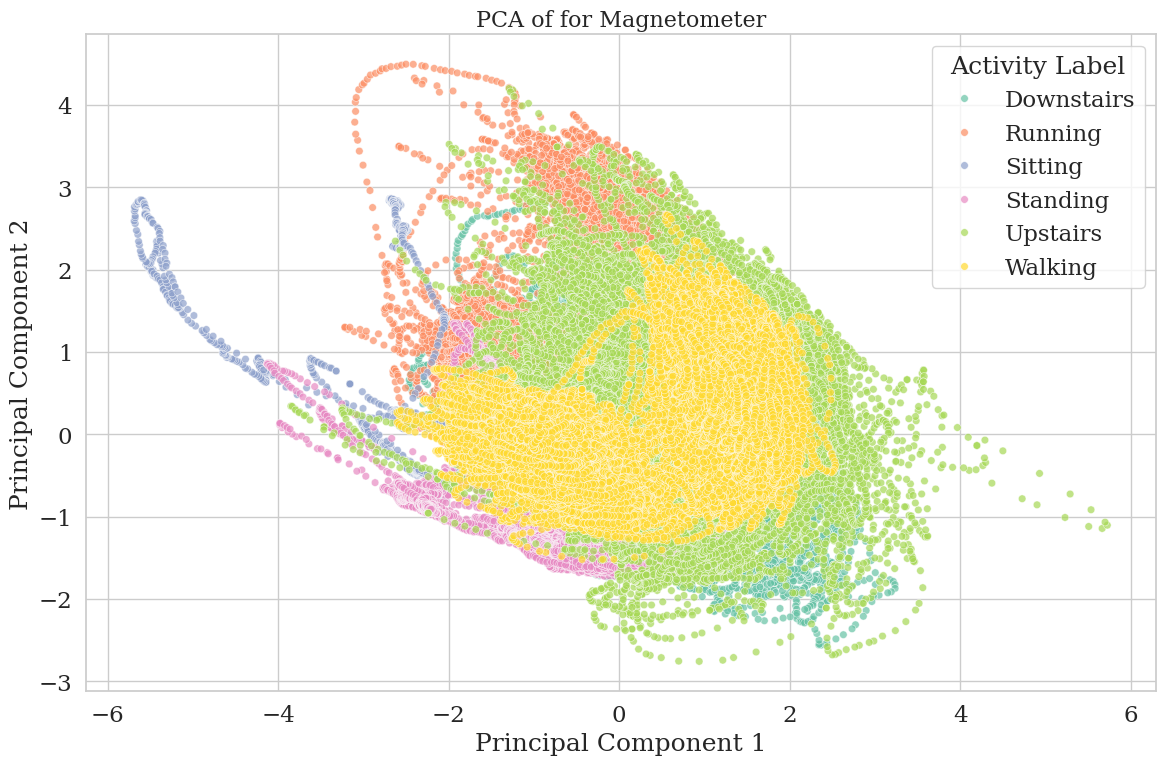} 
        \caption{PCA for Magnetometer}
        \label{pca:sub3}
        
    \end{subfigure}
    \caption{PCA analysis for Shoaib Dataset}
    \label{fig:pca}
\end{figure}

The PCA plots in Fig. \ref{fig:pca} represent the separability of activities in a reduced two-dimensional space for the features of the accelerometer, gyroscope and magnetometer sensor data. Activities such as walking and running form separate clusters for accelerometer data, while stationary activities such as sitting and standing overlap in this data representation. Gyroscope data may present similar patterns of clustering with greater variability. The combined features improve the separability, underlining the merit of multi-sensor data for capturing activity-specific patterns.

\subsection{Data Preprocessing}
For preparing the data, a segmentation approach was used with a sliding window for activity recognition. The designed preprocessing pipeline converts raw time-series sensor data into structured inputs that are suitable for any machine learning model. More precisely, there were nine features around accelerometer readings \( A_x, A_y, A_z \), gyroscope readings \( G_x, G_y, G_z \), and magnetometer readings \( M_x, M_y, M_z \). These were then segmented into fixed-sized windows of \( N_{\text{W}} = 200 \) samples, corresponding to a fixed-length time interval, with a step size of 20 samples to create overlapping windows. Each segment was a multi-dimensional time-series of shape \( (N_{\text{W}} \times N_{\text{F}}) \), where \( N_{\text{F}} = 9 \).

\subsection{Zero-shot capability of LLMs}
We explored Large Language Models' zero-shot capability in human activity recognition tasks to see how the pre-trained LLM can interpret and analyze IMU data, without requiring task-specific fine-tuning.
We experimented with the LLama3-8B model, providing it with specific prompts and statistical features extracted from IMU data to assess its capability in classifying human activities and reasoning through its decisions step-by-step.

\begin{figure}[h!]
    \centering
    \includegraphics[width=\linewidth]{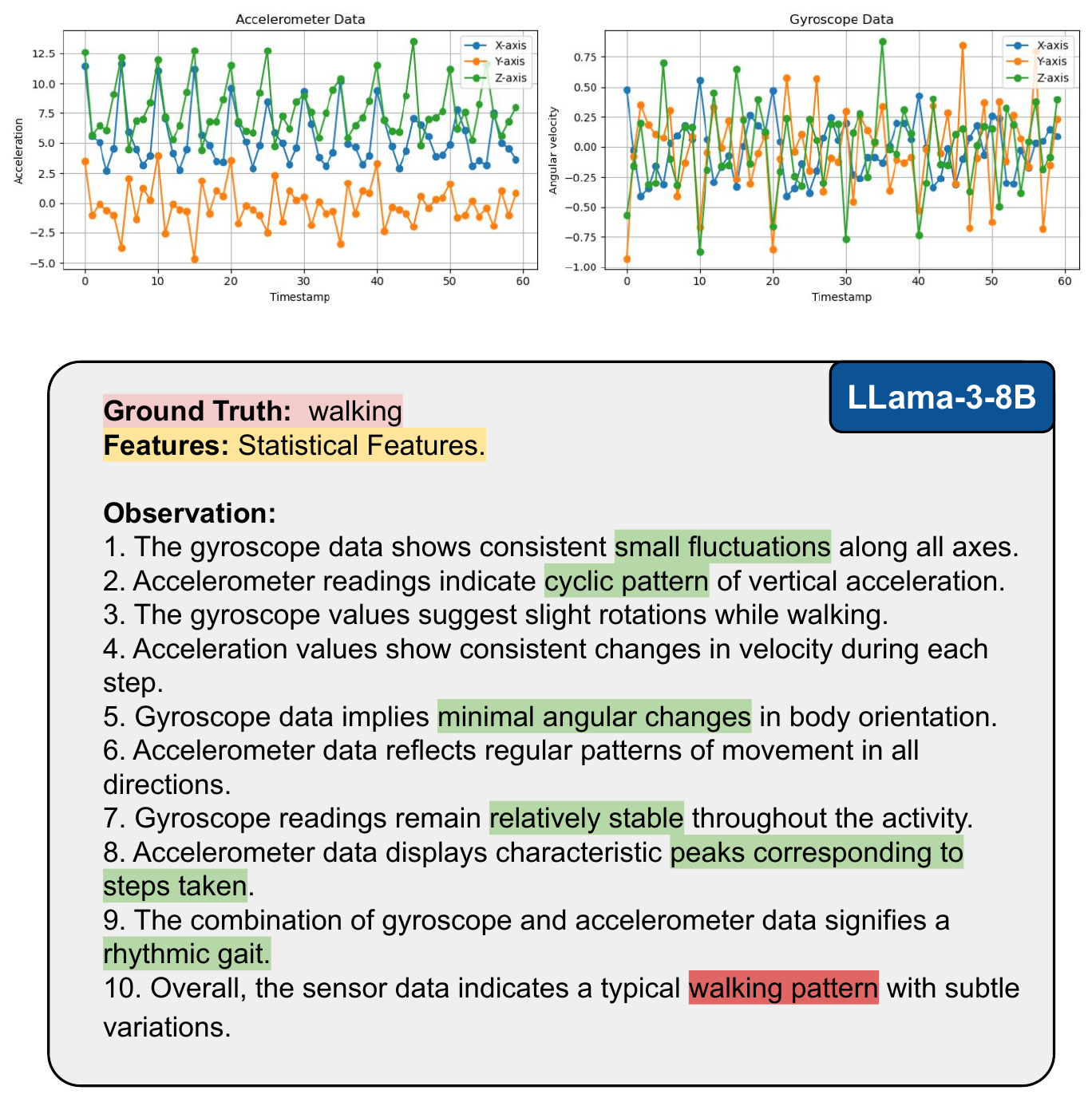} 
    \caption{Zero-shot capability of LLama-3-8b model in HAR}
    \label{fig:zero}
\end{figure}

Figure \ref{fig:zero} demonstrates the zero-shot capability of the LLama-3-8B model in classifying human activity.
Without prior training on this specific dataset or task, the model accurately identifies key patterns such as cyclic variations in acceleration, stable angular changes, and rhythmic gait indicative of walking.

\subsection{Sensor-text modality alighment}
Aligning time-series sensor data with text presents a significant challenge. Unlike modalities such as images or audio, sensor data lacks rich semantics, making it less interpretable for large language models. Previous research on integrating LLMs with Human Activity Recognition (HAR) typically used raw sensor data as input, converting numerical sequences into textual formats before feeding them to the model. While this approach allows LLMs to process sensor data, it comes with significant drawbacks, including high token requirements and computational overhead.
\begin{figure}[h!]
    \centering
    \includegraphics[width=\linewidth]{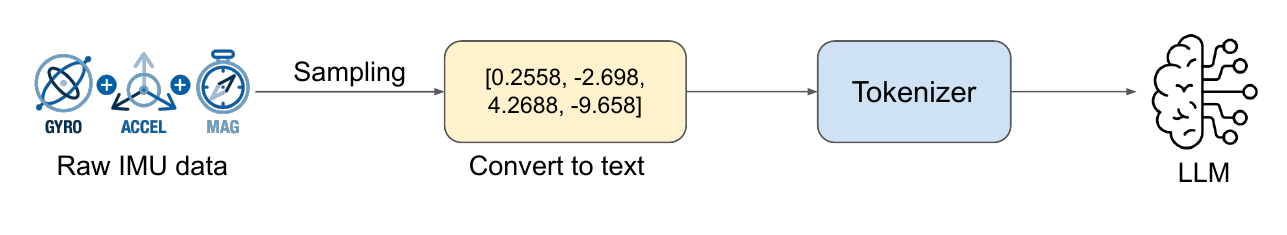} 
    \caption{Traditional approach of align numerical sensor data for LLM input.}
    \label{fig:token}
\end{figure}

\begin{figure}[h!]
    \centering
    \includegraphics[width=\linewidth]{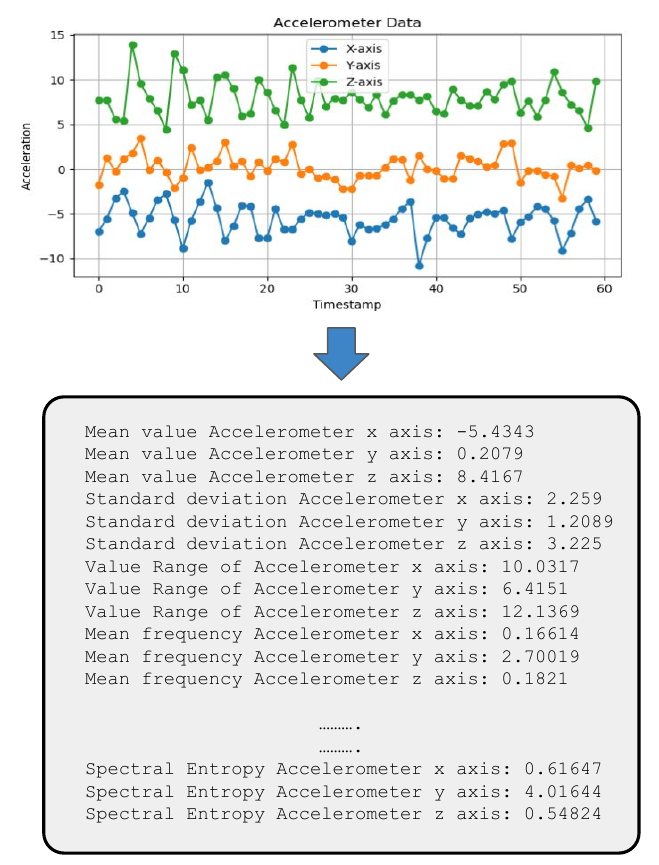} 
    \caption{Statistical Feature Extraction from raw IMU data.}
    \label{fig:feature_extraction}
\end{figure}

Models like LLama-3 8B, which have a maximum token limit of 4096, struggle to handle long sequences of data with high sampling rates. This often leads to truncation, resulting in a loss of valuable information.
To address this issue, we adopted a more efficient approach by extracting statistical features from raw IMU data and using these condensed representations as input to the LLM. 
Figure \ref{fig:feature_extraction} illustrates our feature extraction process from raw IMU data. We extracted both time-domain and frequency-domain features. 
For the time-domain features, we calculated statistical measures such as the mean, standard deviation, and range for all three axes (X, Y, Z) of the accelerometer and gyroscope readings. These features capture the overall variability, central tendency, and extent of the motion data over time.
For the frequency-domain features, we employed techniques such as Fast Fourier Transform (FFT) to analyze the signal's spectral properties. Features like mean frequency, spectral entropy, and band power for high and low-frequency ranges were derived. These metrics provide insights into the periodicity and energy distribution of the motion signals, which are critical for distinguishing between different activities.


\subsection{Fine-Tuning}
For fine-tuning, we utilized Parameter-Efficient Fine-Tuning (PEFT) techniques with Low-Rank Adaptation (LoRA) \cite{hu2021lora} applied to the LLama3-8B model. 
This approach allowed us to efficiently adapt the large-scale pre-trained model to our specific Human Activity Recognition task without requiring a full retraining of all parameters. 
The LoRA configuration included a rank \( r = 128 \), a LoRA alpha value of \( 32 \), and a dropout rate of \( 0.05 \).
The learning rate was set to \( 2 \times 10^{-4} \). 
Inspired by LLaVA \cite{liu2024visual}, we adopted an instruction-tuning format for training, framing the HAR tasks as instruction-following problems. The instruction tuning is completed on a single A-100 GPU.

\begin{figure*}[]
    \centering
    \begin{subfigure}[b]{0.8\linewidth}
        \centering
        \includegraphics[width=\textwidth]{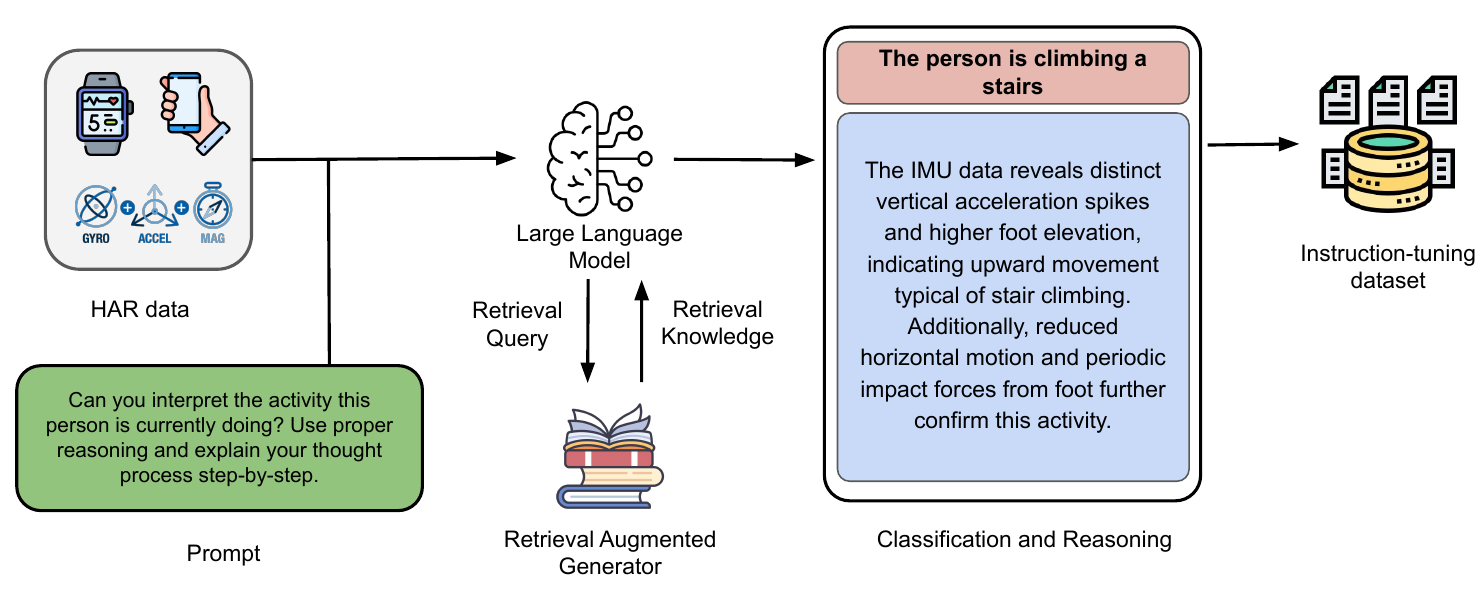} 
        \caption{Creating Instruction Tuning Dataset using RAG}
        \label{fig:sub1}
    \end{subfigure}
    \hfill
    \begin{subfigure}[b]{0.8\linewidth}
        \centering
        \includegraphics[width=\textwidth]{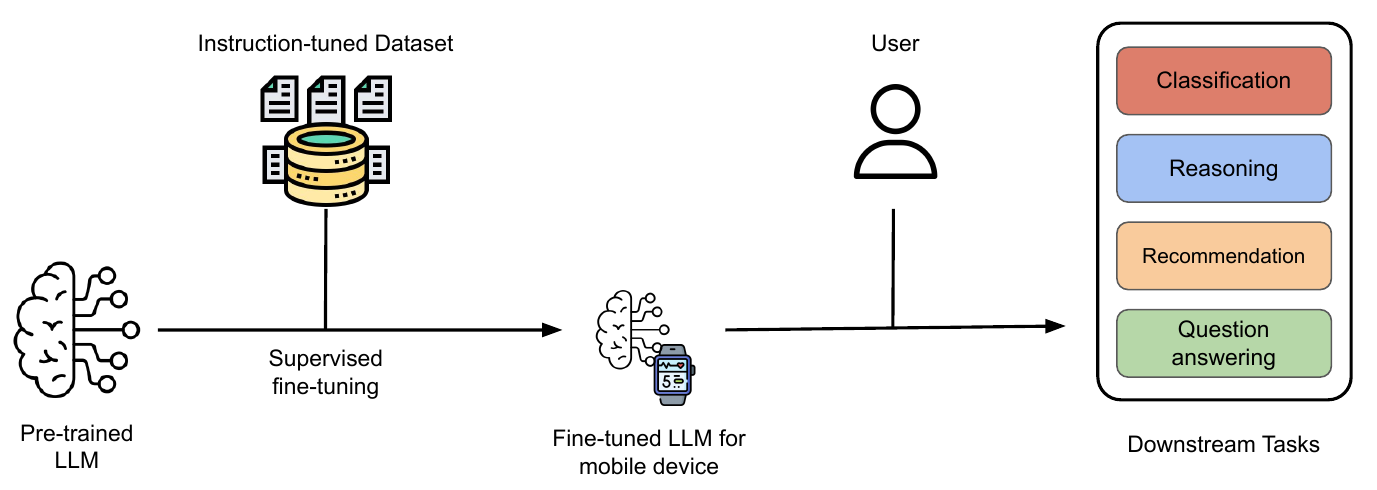} 
        \caption{Fine-tuning LLM for various downstream tasks}
        \label{fig:sub2}
    \end{subfigure}

    \caption{Overview of the proposed approach.}
    \label{fig:fine-tuning}
\end{figure*}



\section{Evaluation}

\subsection{Experiment setup}

\subsubsection{Baseline Models}
For the activity recognition evaluation, we used four baseline models: Support Vector Machine, Random Forest, Deep Neural Network, and Long-Short-Term Memory. SVM and RF are traditional machine learning models where SVM makes use of hyperplanes for classification and RF uses ensemble learning with the help of decision trees. On the other hand, DNN and LSTM are deep learning models. DNN captures complex feature representations using a fully connected layer, while LSTM is a kind of recurrent neural network that was developed to capture sequential data in such a way that its long-term dependencies are maintained. Hence, these models provide an overall framework for the comparison between traditional and deep learning approaches in human activity recognition.

\subsubsection{Evaluation Metrics}

To evaluate the performance of the models, we employed four metrics: Accuracy, Precision, Recall, and F1 Score. 

\textbf{Accuracy:} Accuracy is defined as the ratio of correctly predicted instances to the total number of instances. It measures the overall effectiveness of the model and is calculated as:
\[
\text{Accuracy} = \frac{\text{TP} + \text{TN}}{\text{TP} + \text{TN} + \text{FP} + \text{FN}}
\]
where \textbf{TP} represents True Positives, \textbf{TN} represents True Negatives, \textbf{FP} represents False Positives, and \textbf{FN} represents False Negatives.

\textbf{Precision:} Precision quantifies the proportion of correctly predicted positive instances out of all instances predicted as positive. It evaluates the model's ability to avoid \textbf{FP} and is expressed as:
\[
\text{Precision} = \frac{\text{TP}}{\text{TP} + \text{FP}}
\]

\textbf{Recall:} Recall, also known as sensitivity or true positive rate, measures the proportion of actual positives correctly identified by the model. It emphasizes the model's ability to capture \textbf{TP} and is calculated as:
\[
\text{Recall} = \frac{\text{TP}}{\text{TP} + \text{FN}}
\]

\textbf{F1 Score:} The F1 score is the harmonic mean of Precision and Recall, providing a balanced measure that accounts for both \textbf{FP} and \textbf{FN}. It is particularly useful for imbalanced datasets and is given by:
\[
\text{F1 Score} = 2 \times \frac{\text{Precision} \times \text{Recall}}{\text{Precision} + \text{Recall}}
\]

These metrics collectively offer a detailed evaluation of model performance, enabling a thorough comparison of classification models across multiple datasets.

\subsection{Experiment results}
First, we evaluate the performance of traditional HAR classification techniques on each dataset.

\begin{figure*}[h]
    \centering
    \begin{subfigure}[b]{0.22\textwidth}
        \centering
        \includegraphics[width=\textwidth]{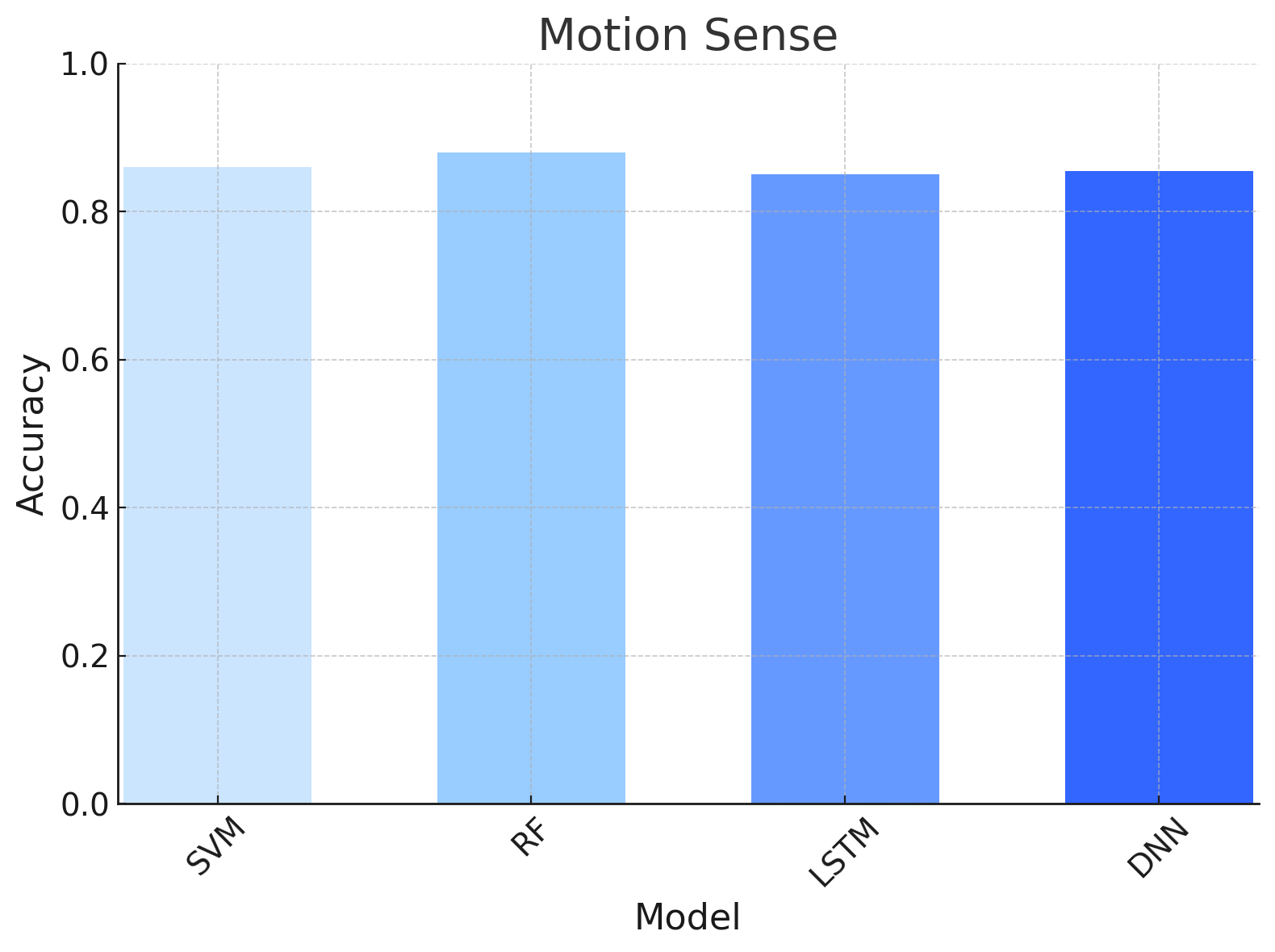} 
        \caption{MotionSense}
        \label{perf:sub1}
    \end{subfigure}
    \hfill
    \begin{subfigure}[b]{0.22\textwidth}
        \centering
        \includegraphics[width=\textwidth]{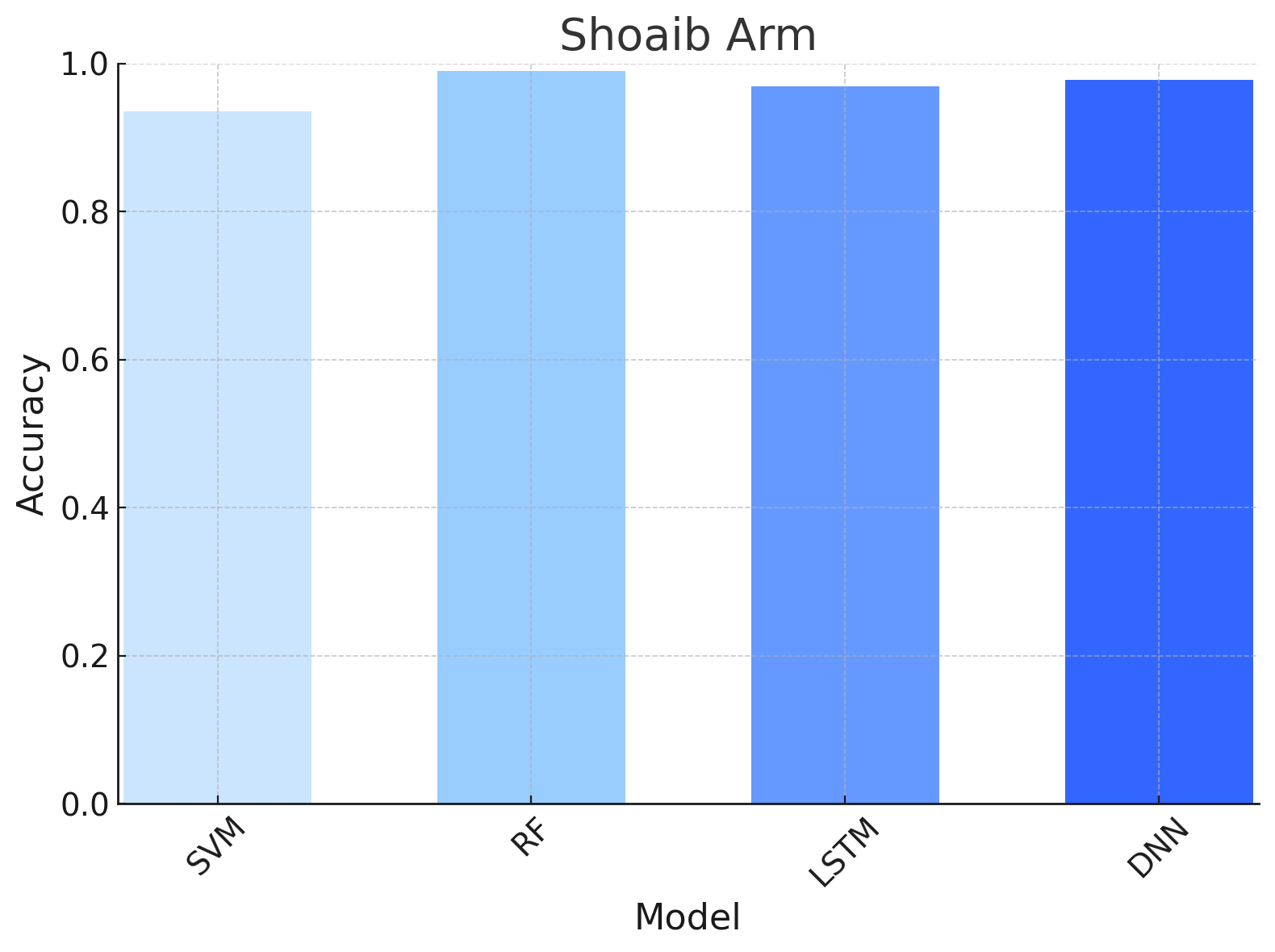} 
        \caption{Shoaib}
        \label{perf:sub2}
    \end{subfigure}
    \hfill
    \begin{subfigure}[b]{0.22\textwidth}
        \centering
        \includegraphics[width=\textwidth]{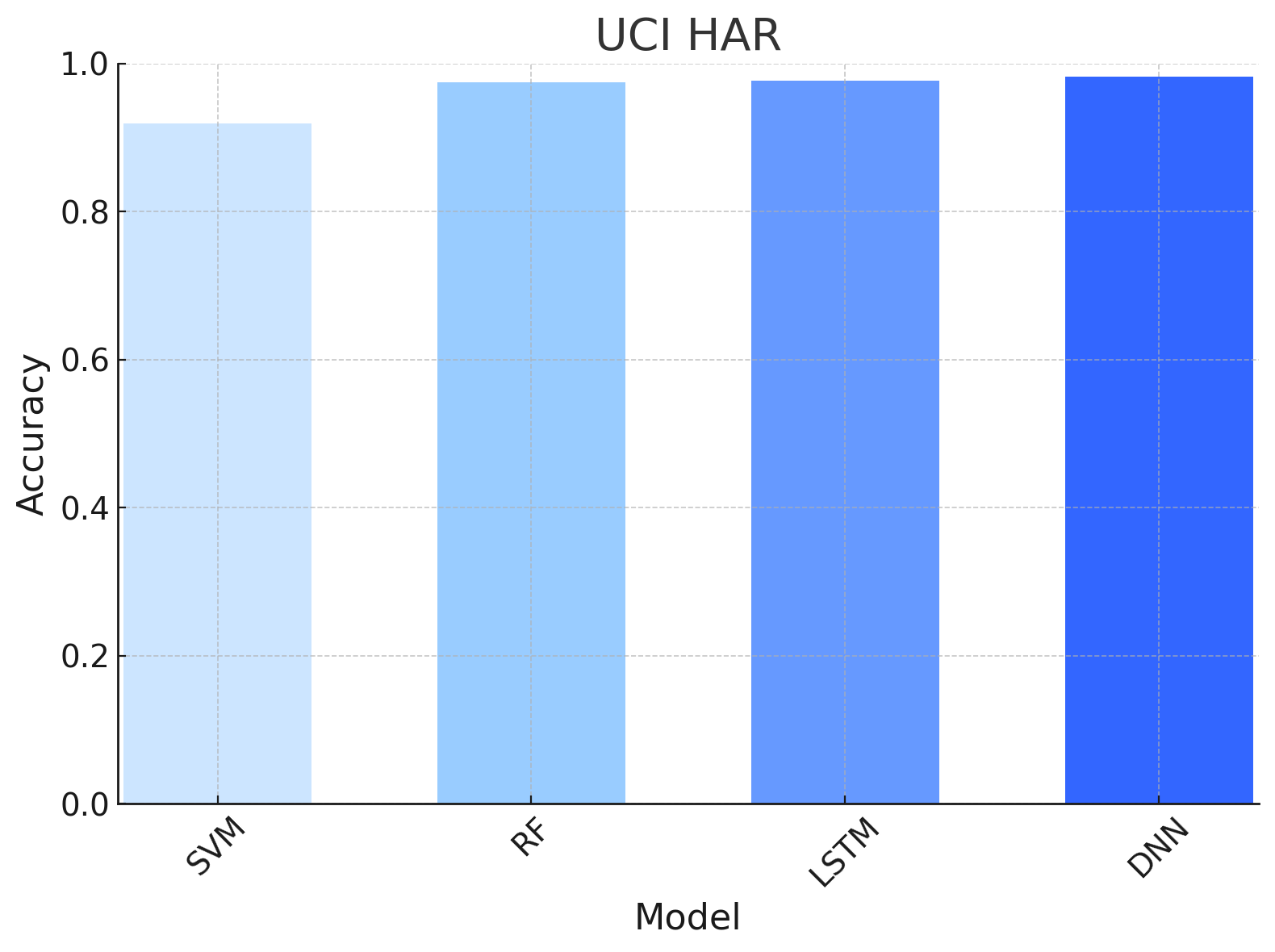} 
        \caption{UCI HAR}
        \label{perf:sub3}
    \end{subfigure}
    \hfill
    \begin{subfigure}[b]{0.22\textwidth}
        \centering
        \includegraphics[width=\textwidth]{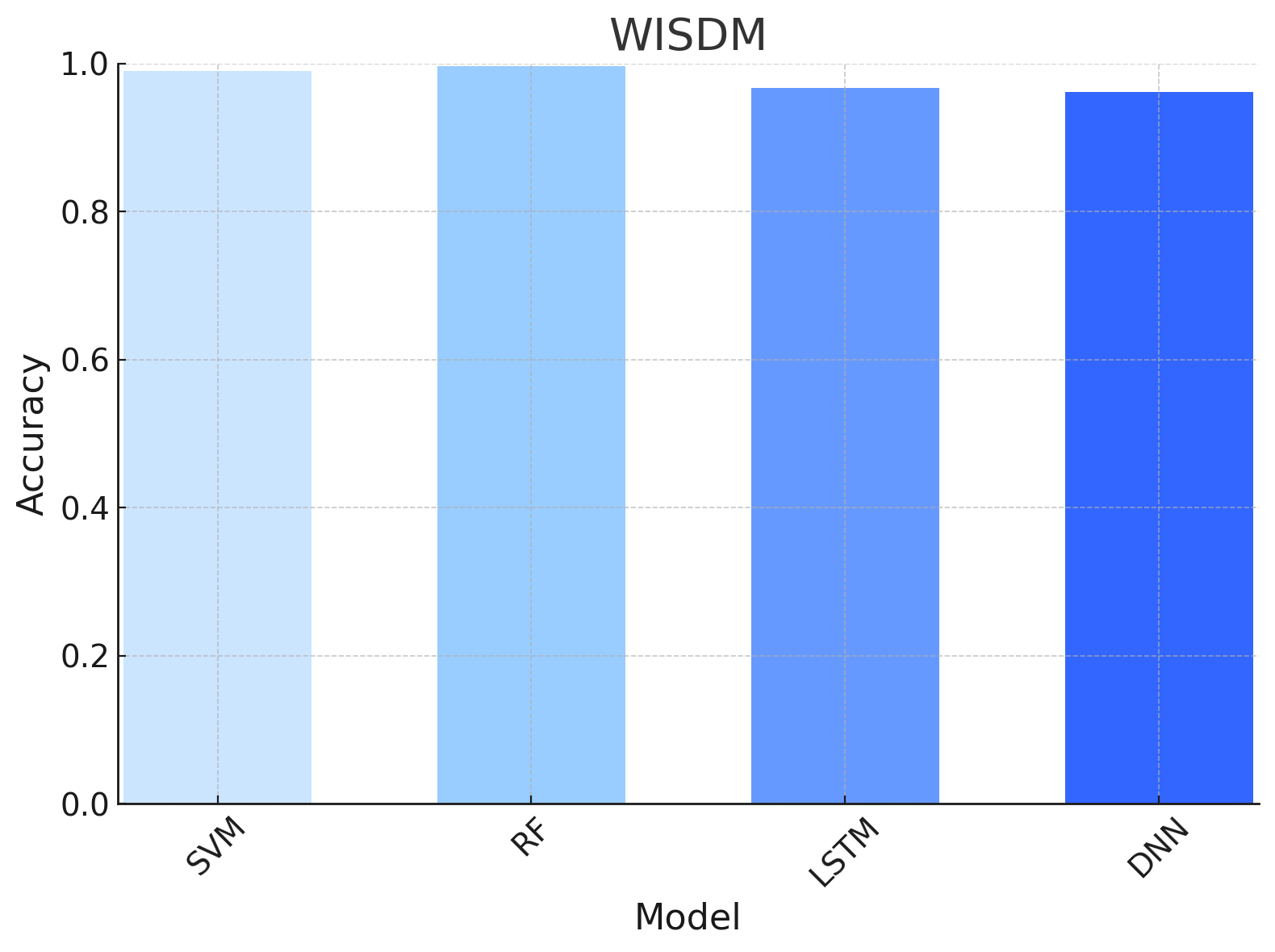} 
        \caption{WISDM}
        \label{perf:sub4}
    \end{subfigure}
    
    \caption{Performance analysis for four datasets.}
    \label{fig:perf}
\end{figure*}

\begin{table}[ht]
\centering
\renewcommand{\arraystretch}{1.3} 
\resizebox{\columnwidth}{!}{%
\begin{tabular}{l l c c c c}
\toprule
\textbf{Dataset} & \textbf{Model Name} & \textbf{Accuracy} & \textbf{Macro Precision} & \textbf{Macro Recall} & \textbf{Macro F1} \\
\midrule
Shoaib Arm & SVM  & 0.9352 & 0.9365 & 0.9213 & 0.9270 \\
           & RF   & 0.9894 & 0.9892 & 0.9867 & 0.9879 \\
           & LSTM & 0.9695 & 0.9664 & 0.9645 & 0.9645 \\
           & DNN  & 0.9782 & 0.9765 & 0.9753 & 0.9758 \\
UCI HAR    & SVM  & 0.9190 & 0.9148 & 0.9169 & 0.9144 \\
           & RF   & 0.9747 & 0.9747 & 0.9692 & 0.9716 \\
           & LSTM & 0.9765 & 0.9716 & 0.9730 & 0.9719 \\
           & DNN  & 0.9821 & 0.9796 & 0.9806 & 0.9800 \\
WISDM      & SVM  & 0.9895 & 0.9754 & 0.9589 & 0.9656 \\
           & RF   & 0.9965 & 0.9982 & 0.9833 & 0.9904 \\
           & LSTM & 0.9668 & 0.8317 & 0.7628 & 0.7807 \\
           & DNN  & 0.9616 & 0.9215 & 0.8136 & 0.7894 \\
Motion Sense & SVM  & 0.8603 & 0.8390 & 0.8116 & 0.8203 \\
           & RF   & 0.8796 & 0.8590 & 0.8410 & 0.8466 \\
           & LSTM & 0.8508 & 0.8181 & 0.8081 & 0.8110 \\
           & DNN  & 0.8544 & 0.8206 & 0.8185 & 0.8192 \\
\bottomrule
\end{tabular}%
}
\caption{Classification Metrics Summary}
\label{tab:metrics-summary}
\end{table}

\begin{figure*}[h]
    \centering
    \includegraphics[width=0.9\textwidth]{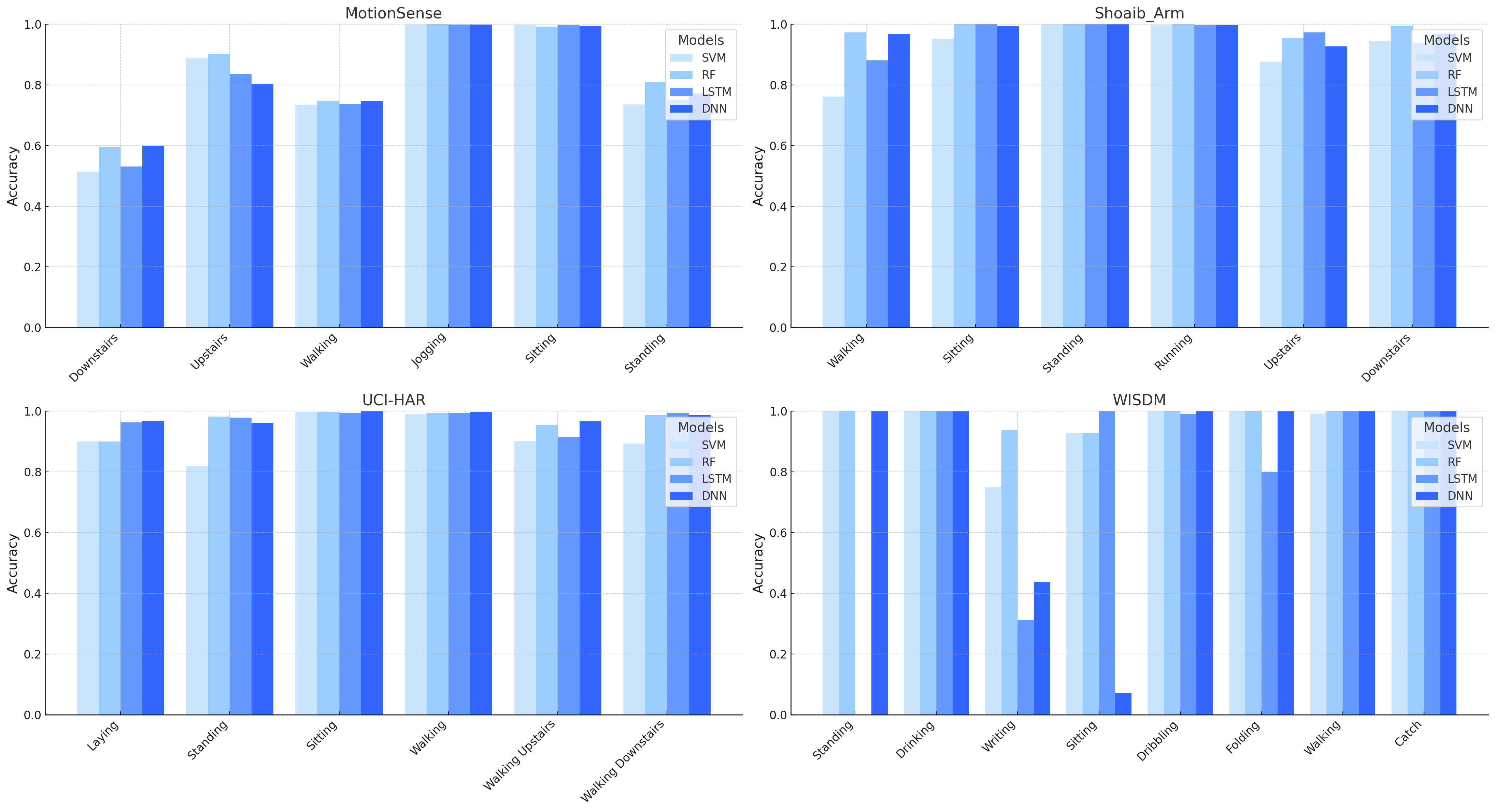} 
    \caption{Classwise performance analysis for four datasets.}
    \label{classwise}
\end{figure*}

Fig. \ref{fig:perf} illustrates the performance comparison of Macro F1 scores for SVM, RF, LSTM, and DNN on Motion Sense, Shoaib, UCI HAR, and WISDM datasets. Overall, Random Forest produces the highest performance in most datasets, thus proving its strength in robustness and adaptability. The DNN and LSTM deep learning models present a very strong performance, especially in those datasets that have a rich temporal pattern, such as UCI HAR. However, they are limited in datasets with high variability, such as WISDM. SVM performs reliably but lags behind RF and DNN, reflecting its restricted ability to capture complex relationships. These results emphasize the importance of tailoring model selection to dataset characteristics to maximize performance.

Table \ref{tab:metrics-summary} presents the performance results for SVM, RF, LSTM, and DNN in classification tasks with four different datasets. The comparison metrics used were Accuracy, Macro Precision, Macro Recall, and Macro F1 Score. This comparison revealed differences in both strengths and weaknesses within various combinations of model-dataset pairs.

For the \textbf{Shoaib Arm Dataset}, we got the accuracy of RF as  
0.9894 and Macro F1 Score of 0.9879. LSTM follows suit with an accuracy value of 0.9695 and DNN follows closely with an accuracy value of 0.9782, while SVM performs at 0.9352, ranking lower among these models. 
%
In the case of the \textbf{UCI HAR Dataset}, DNN achieves the best overall performance with an accuracy of 0.9821 and a Macro F1 Score of 0.9800. LSTM and RF also had good performances with an accuracy of 0.9765 and 0.9747, respectively. SVM is behind them with an accuracy of 0.9190. The high scores of DNN and LSTM suggest that deep learning models effectively capture the temporal patterns in this dataset.

There is a big variation in the performance of RF for the \textbf{WISDM Dataset}, with near-perfect results-0.9965 for accuracy and 0.9904 for Macro F1 Score-always outperforming all the others. It follows that SVM also runs smoothly on this dataset with 0.9895 accuracy and 0.9656 Macro F1 Score. The deep learning models, however, show a remarkable fall, especially LSTM, which produced 0.9668 for accuracy and 0.7807 for Macro F1 values. This suggests that the dataset's properties may pose challenges for the sequential models.

In the \textbf{Motion Sense Dataset}, traditional machine learning models such as SVM and RF perform slightly better than deep learning algorithms. RF yielded the highest accuracy value of 0.8796 with a Macro F1 Score of 0.8466, while SVM performed second with an accuracy value of 0.8603. LSTM and DNN perform similarly, their accuracy values being 0.8508 and 0.8544, respectively. The relatively low scores compared to other datasets hint at higher variability in data or imbalance in activity classes.

RF consistently outperforms other models across all datasets, demonstrating its robustness for activity recognition tasks. Deep learning models (DNN and LSTM) excel in datasets with strong temporal patterns, such as UCI HAR, but show limited performance on datasets like WISDM. SVM, while effective, generally underperforms compared to RF and DNN, indicating its limitations in capturing complex patterns. Results emphasize the selection of a model that best suits the nature of the dataset for optimal performance.

Fig. \ref{classwise} shows the class-wise performance of four models, SVM, RF, LSTM, and DNN, across four datasets, namely MotionSense, Shoaib, UCI HAR, and WISDM. Overall, RF tends to show pretty consistent and highest performance in most of the activities across datasets, specifically doing exceptionally well on structured and stationary activities like \textit{Standing} and \textit{Sitting} for MotionSense, Shoaib, and UCI HAR datasets. DNN also shows competitive accuracy, especially for datasets with obvious temporal patterns, such as UCI HAR. On the other hand, LSTM performs variably: it performs well on dynamic activities such as \textit{Jogging} but does poorly on diverse and complex activities in the WISDM dataset, such as \textit{Dribbling} and \textit{Writing}. SVM is generally reliable but usually behind RF and DNN, especially for harder classes. The analysis shows that RF is relatively robust for activity classification tasks and the selection of models should be based on data characteristics and activities.

\begin{figure}[h]
    \centering
    \begin{subfigure}[b]{0.48\columnwidth}
        \centering
        \includegraphics[width=\textwidth]{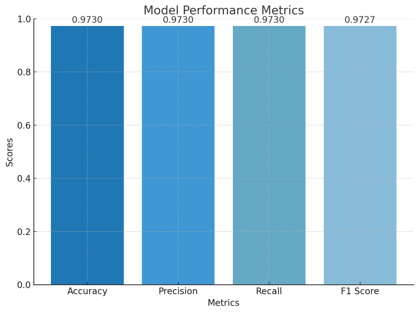} 
        \caption{Same-Modality}
        \label{modality:sub1}
    \end{subfigure}
    \hfill
    \begin{subfigure}[b]{0.48\columnwidth}
        \centering
        \includegraphics[width=\textwidth]{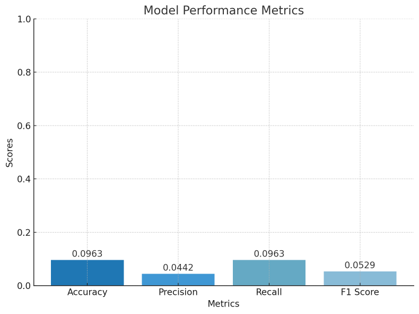} 
        \caption{Cross-Modality}
        \label{modality:sub2}
    \end{subfigure}
    
    \caption{Comparison of Same-Modality vs. Cross-Modality Performance}
    \label{fig:modality}
\end{figure}

These bar charts present some model performance metrics such as accuracy, precision, recall, and F1 score when trained and tested on the same dataset versus on different ones. As expected, the same-modality performances are decidedly higher across the board, with approximately 0.973 for accuracy, precision, recall, and F1 score in that order. This means that the model generalizes well for the same dataset since there are similar feature distributions and patterns. In contrast, the results in the cross-modality setting show significant decreases in performance for all metrics: Accuracy and recall go down to about 0.0963, while Precision and F1 Score decrease further to 0.0442 and 0.0529, respectively.  This performance gap underlines the challenge of cross-modality activity recognition and highlights the importance of domain adaptation techniques to improve generalizability across data sets. 

\begin{table}[h]
\renewcommand{\arraystretch}{1.7}
\resizebox{\columnwidth}{!}{%
\begin{tabular}{lllllll}
\toprule
\multicolumn{1}{c}{Dataset}        & \multicolumn{2}{c}{\textbf{HHAR}}                               & \multicolumn{2}{c}{\textbf{MotionSense}}                        & \multicolumn{2}{c}{\textbf{Shoaib}}                            \\ \midrule
\multicolumn{1}{c|}{Test Subject}  & \multicolumn{1}{c}{Seen} & \multicolumn{1}{c|}{Unseen} & \multicolumn{1}{c}{Seen} & \multicolumn{1}{c|}{Unseen} & \multicolumn{1}{c}{Seen} & \multicolumn{1}{c}{Unseen} \\ \hline
\multicolumn{1}{l|}{Random Forest} & 0.97                     & \multicolumn{1}{l|}{0.67}   & \textbf{0.84}                     & \multicolumn{1}{l|}{0.58}   & \textbf{0.98}                     & 0.67                       \\
\multicolumn{1}{l|}{SVM}           & 0.91                     & \multicolumn{1}{l|}{0.47}   & 0.82                     & \multicolumn{1}{l|}{0.42}   & 0.92                     & 0.56                       \\
\multicolumn{1}{l|}{DNN}           & \textbf{0.98}                     & \multicolumn{1}{l|}{0.61}   & 0.81                     & \multicolumn{1}{l|}{0.38}   & 0.97                     & 0.64                       \\ \hline
\multicolumn{1}{l|}{LLama-3-8B }          & -                     & \multicolumn{1}{l|}{0.65}   & -                     & \multicolumn{1}{l|}{0.61}   & -                     & 0.67       \\
\multicolumn{1}{l|}{LLama-3-8B (Fine-tuned)}          & 0.83                     & \multicolumn{1}{l|}{\textbf{0.75}}   & 0.77                     & \multicolumn{1}{l|}{\textbf{0.65}}   & 0.79                     & \textbf{0.71}       \\ \bottomrule               
\end{tabular}
}
\caption{Classification Result}
\label{tab:comparison-table}
\end{table}

\begin{figure}[h!]
    \centering
    \begin{subfigure}[b]{\columnwidth}
        \centering
        \includegraphics[width=\linewidth]{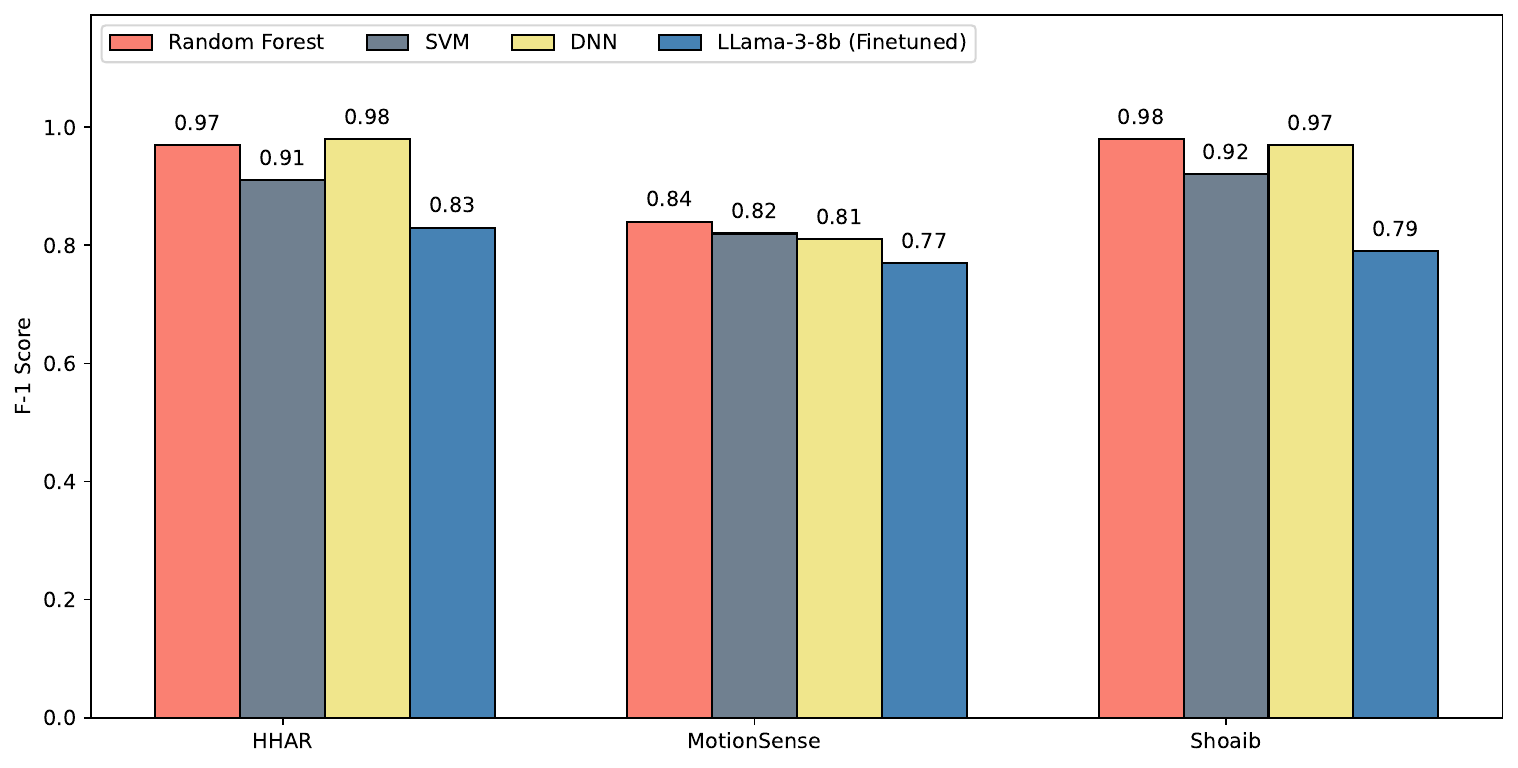} 
        \caption{Classification result on seen test dataset}
        \label{fig:sub1}
    \end{subfigure}
    \hfill
    \begin{subfigure}[b]{\columnwidth}
        \centering
        \includegraphics[width=\linewidth]{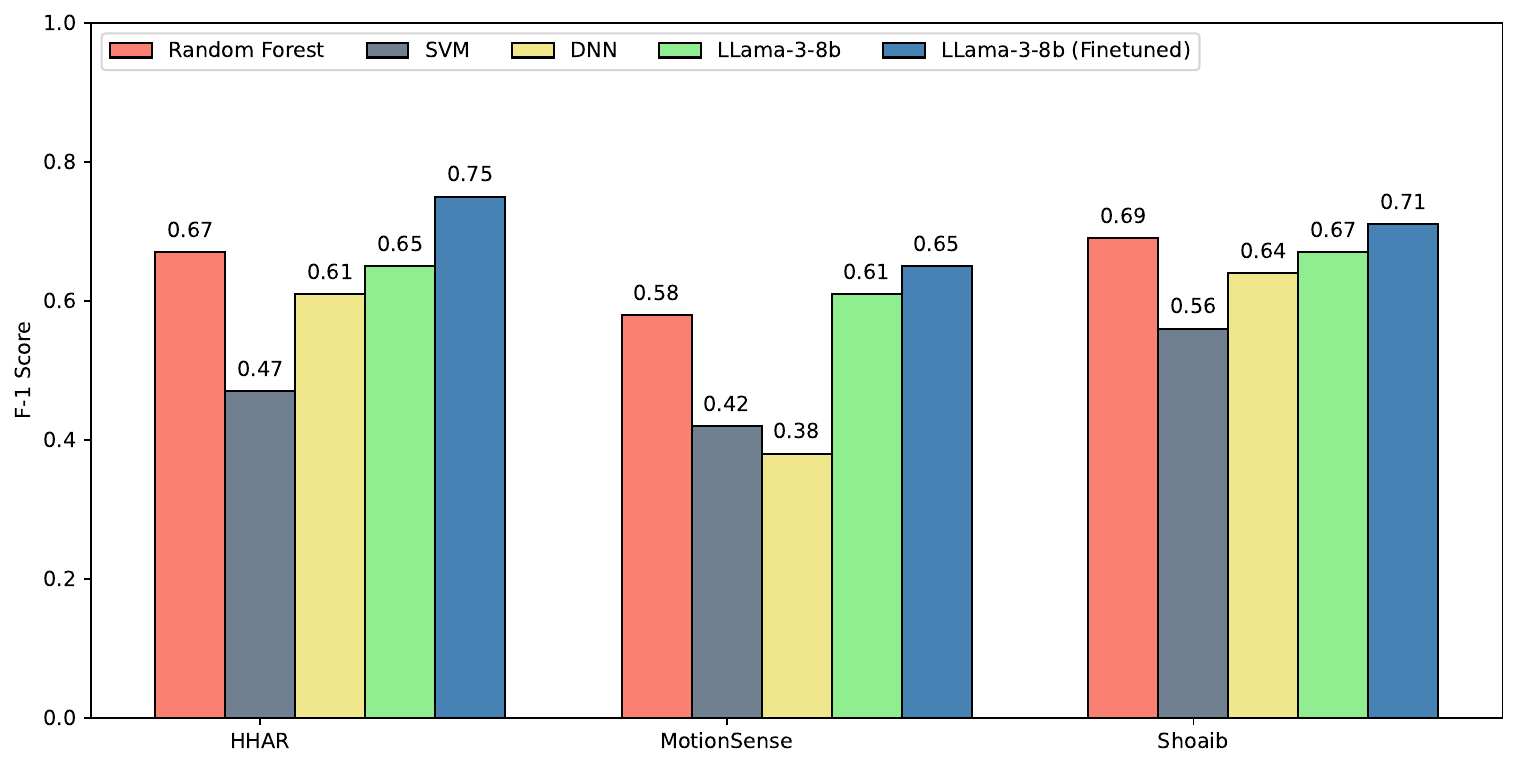} 
        \caption{Classification result on unseen test dataset}
        \label{fig:sub2}
    \end{subfigure}
    
    \caption{Performance comparison.}
    \label{fig:llama-compare-barplot}
\end{figure}

Table \ref{tab:comparison-table} and Figure \ref{fig:llama-compare-barplot} provides a performance comparison of various models—Random Forest, Support Vector Machine (SVM), and Deep Neural Network (DNN)—against our proposed framework for a classification task across three datasets: HHAR, MotionSense, and Shoaib. The experiments were conducted under two distinct settings: 'seen' and 'unseen'. In the 'seen' setting, the models were trained and evaluated on the same data distribution, whereas in the 'unseen' setting, the models encountered a different data distribution during evaluation, which was not part of their training data. 
From Table \ref{tab:comparison-table}, we observe that traditional approaches like RF and DNN perform better in the 'seen' case, whereas the LLM-based approach excels in the 'unseen' case. This is likely because traditional models are optimized for recognizing patterns in specific data distributions they were trained on, making them highly effective when the test data closely resembles the training IMU dataset. In contrast, LLM-based methods, leverage their inherent ability to generalize across diverse and unseen IMU data distributions.

\begin{figure}[h]
    \centering
    \includegraphics[width=\linewidth]{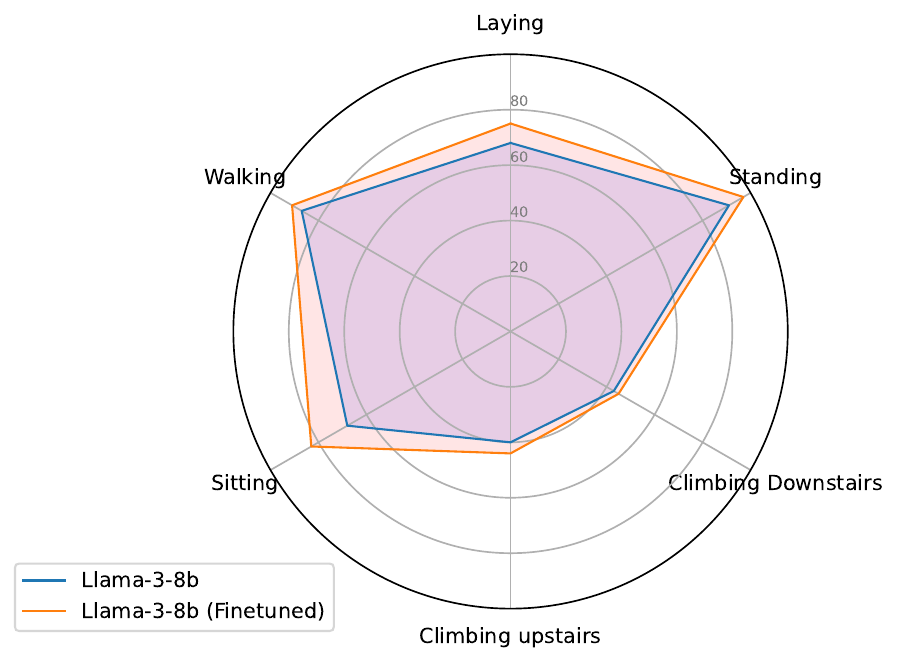} 
    \caption{Classwise performance analysis on HHAR datasets.}
    \label{radar_plot}
\end{figure}

Figure \ref{radar_plot} presents a radar chart comparing the class-wise performance of the Llama-3-8b model and its finetuned counterpart on the HHAR dataset. Activities such as Walking and Standing demonstrate higher accuracy, while activities like Climbing Upstairs and Climbing Downstairs tend to have lower accuracy.
The higher accuracy for Walking and Standing can be attributed to their distinctive and consistent patterns in sensor data, which are easier for the model to identify and classify. These activities often involve repetitive and predictable movements, leading to less variability within the same class. In contrast, activities like Climbing Stairs involve more complex and variable motion patterns, which can overlap with other classes, making them harder to distinguish accurately. 
The finetuned Llama-3-8b model outperforms the base model across most activity classes, with notable improvements in Laying, and Sitting.

\section{Issues encountered}

In the process of integrating a large language model for the human activity recognition task, several issues were identified that impacted both the development and deployment phases.

\subsection{Model Limitations}
While LLMs demonstrate robust natural language processing capabilities, they often struggle to interpret raw IMU sensor data. Unlike textual data, IMU data requires sophisticated preprocessing to extract meaningful features, which LLMs are not inherently designed to handle.
A significant issue observed in LLMs, particularly smaller models such as LLama-3-8B and LLama-3-1B, is their tendency to hallucinate. These models can generate made-up reasoning or predictions that deviate significantly from true activity patterns. 
Larger models like GPT-4 exhibit better performance compared to smaller, open-source models. However, their substantial size and resource requirements pose significant challenges for on-device implementation.

\subsection{Modality Alignment}
LLMs excel in processing text-based data, leveraging their pretraining on vast corpora of textual information to generate outputs. IMU data, on the other hand, is inherently time-series in nature and consists of numerical sequences captured over time. Aligning such data with LLMs, which are primarily designed for textual input, introduces a key challenge. In this work, we overcome this challenge by extracting statistical features from the raw IMU data and transforming them into a structured format compatible with text-based representation.

\section{Discussion and Future work}

In this work, we conducted experiments to evaluate the performance of large language models in human activity recognition tasks. We explored the challenges associated with utilizing LLMs for HAR tasks and proposed a solution to address these limitations. Our study includes an investigation into the zero-shot capabilities of LLMs for HAR, examining how LLMs can reason about their decisions and infer patterns from complex, multimodal data. Additionally, we developed a fine-tuning pipeline to tailor LLMs for more specific HAR-related tasks.

To assess the effectiveness of LLMs, we compared their performance with traditional machine learning (ML) and deep learning (DL) approaches in both seen and unseen data settings. Our results demonstrate that traditional methods outperform LLMs when trained on data from the same distribution, owing to their ability to specialize in specific patterns within a controlled dataset. However, in scenarios involving unseen data, LLMs show superior performance. This advantage arises because LLMs leverage their pretraining on diverse and extensive datasets, enabling them to generalize better and make informed assumptions about novel activities, even without explicit prior training on similar examples.

We addressed the critical challenges of integrating sensor data with large language models (LLMs) and proposed a method to bridge the gap between these modalities. Recognizing the limitations of manual feature extraction, future work will focus on leveraging pre-trained encoders\cite{yuan2024self} to automatically extract features from sensor data. By aligning these encoder-driven features with LLMs, inspired by successful integration techniques in vision and audio domains\cite{gardner2023llark}, we aim to enhance the performance and scalability of our framework. This approach has the potential to streamline the feature extraction process, improve accuracy, and enable more seamless multimodal integration. Ultimately, these advancements will further the applicability of LLMs in human activity recognition and beyond, paving the way for robust, efficient, and interpretable AI-driven solutions.



\section{Conclusion}
This paper bridges the gap between IMU sensory data and large language models, presenting a novel approach that extends beyond traditional classification tasks to include report generation and reasoning capabilities. By addressing the inherent challenges of modality alignment between time-series IMU data and Language models, we proposed a framework that extracts meaningful statistical features from raw sensor data and integrates them with the reasoning power of LLMs. This approach enables LLMs to not only classify activities but also provide explainable insights and generate detailed analyses based on their predictions.
Our study demonstrates the potential of LLMs to excel in zero-shot and unseen data scenarios, where their ability to generalize from extensive pretraining proves advantageous. By developing a fine-tuning pipeline, we enhanced the performance of LLMs for HAR-related tasks. The comparative analysis with traditional ML and DL methods highlights the complementary strengths of these approaches—traditional models excel in distribution-specific scenarios, while LLMs shine in environments requiring generalization and adaptability.
While the integration of LLMs in HAR shows promising potential, particularly in handling unseen data distributions, there is still a need to improve accuracy for broader adoption. Future work could focus on leveraging feature encoders to better extract and represent key characteristics from raw sensor data, enhancing the alignment between modalities and boosting overall performance in both seen and unseen scenarios.


\bibliographystyle{ieee_fullname}
\bibliography{references,zhu, song}

@article{li2024personal,
  title={Personal llm agents: Insights and survey about the capability, efficiency and security},
  author={Li, Yuanchun and Wen, Hao and Wang, Weijun and Li, Xiangyu and Yuan, Yizhen and Liu, Guohong and Liu, Jiacheng and Xu, Wenxing and Wang, Xiang and Sun, Yi and others},
  journal={arXiv preprint arXiv:2401.05459},
  year={2024}
}

@Article{informatics5020027,
AUTHOR = {Twomey, Niall and Diethe, Tom and Fafoutis, Xenofon and Elsts, Atis and McConville, Ryan and Flach, Peter and Craddock, Ian},
TITLE = {A Comprehensive Study of Activity Recognition Using Accelerometers},
JOURNAL = {Informatics},
VOLUME = {5},
YEAR = {2018},
NUMBER = {2},
ARTICLE-NUMBER = {27},
URL = {https://www.mdpi.com/2227-9709/5/2/27},
ISSN = {2227-9709},
DOI = {10.3390/informatics5020027}
}

@article{bulling2014tutorial,
  title={A tutorial on human activity recognition using body-worn inertial sensors},
  author={Bulling, Andreas and Blanke, Ulf and Schiele, Bernt},
  journal={ACM Computing Surveys (CSUR)},
  volume={46},
  number={3},
  pages={1--33},
  year={2014},
  publisher={ACM New York, NY, USA}
}

@inproceedings{yang2015deep,
  title={Deep convolutional neural networks on multichannel time series for human activity recognition.},
  author={Yang, Jianbo and Nguyen, Minh Nhut and San, Phyo Phyo and Li, Xiaoli and Krishnaswamy, Shonali},
  booktitle={Ijcai},
  volume={15},
  pages={3995--4001},
  year={2015},
  organization={Buenos Aires, Argentina}
}

@article{hammerla2016deep,
  title={Deep, convolutional, and recurrent models for human activity recognition using wearables},
  author={Hammerla, Nils Y and Halloran, Shane and Pl{\"o}tz, Thomas},
  journal={arXiv preprint arXiv:1604.08880},
  year={2016}
}

@article{jin2023time,
  title={Time-llm: Time series forecasting by reprogramming large language models},
  author={Jin, Ming and Wang, Shiyu and Ma, Lintao and Chu, Zhixuan and Zhang, James Y and Shi, Xiaoming and Chen, Pin-Yu and Liang, Yuxuan and Li, Yuan-Fang and Pan, Shirui and others},
  journal={arXiv preprint arXiv:2310.01728},
  year={2023}
}

@article{Attal2015Dec,
	author = {Attal, Ferhat and Mohammed, Samer and Dedabrishvili, Mariam and Chamroukhi, Faicel and Oukhellou, Latifa and Amirat, Yacine},
	title = {{Physical Human Activity Recognition Using Wearable Sensors}},
	journal = {Sensors},
	volume = {15},
	number = {12},
	pages = {31314--31338},
	year = {2015},
	month = dec,
	issn = {1424-8220},
	publisher = {Multidisciplinary Digital Publishing Institute},
	doi = {10.3390/s151229858}
}

@inproceedings{stisen2015smart,
  title={Smart devices are different: Assessing and mitigatingmobile sensing heterogeneities for activity recognition},
  author={Stisen, Allan and Blunck, Henrik and Bhattacharya, Sourav and Prentow, Thor Siiger and Kj{\ae}rgaard, Mikkel Baun and Dey, Anind and Sonne, Tobias and Jensen, Mads M{\o}ller},
  booktitle={Proceedings of the 13th ACM conference on embedded networked sensor systems},
  pages={127--140},
  year={2015}
}

@article{shoaib2014fusion,
  title={Fusion of smartphone motion sensors for physical activity recognition},
  author={Shoaib, Muhammad and Bosch, Stephan and Incel, Ozlem Durmaz and Scholten, Hans and Havinga, Paul JM},
  journal={Sensors},
  volume={14},
  number={6},
  pages={10146--10176},
  year={2014},
  publisher={Multidisciplinary Digital Publishing Institute}
}

@inproceedings{anguita2013public,
  title={A public domain dataset for human activity recognition using smartphones.},
  author={Anguita, Davide and Ghio, Alessandro and Oneto, Luca and Parra, Xavier and Reyes-Ortiz, Jorge Luis and others},
  booktitle={Esann},
  volume={3},
  pages={3},
  year={2013}
}

@inproceedings{malekzadeh2019mobile,
  title={Mobile sensor data anonymization},
  author={Malekzadeh, Mohammad and Clegg, Richard G and Cavallaro, Andrea and Haddadi, Hamed},
  booktitle={Proceedings of the international conference on internet of things design and implementation},
  pages={49--58},
  year={2019}
}

@misc{wisdm_smartphone,
  author       = {Weiss, Gary},
  title        = {{WISDM Smartphone and Smartwatch Activity and Biometrics Dataset }},
  year         = {2019},
  howpublished = {UCI Machine Learning Repository},
  note         = {{DOI}: https://doi.org/10.24432/C5HK59}
}

@article{AIOT_survey,
author = {Siam, Shakhrul Iman and Ahn, Hyunho and Liu, Li and Alam, Samiul and Shen, Hui and Cao, Zhichao and Shroff, Ness and Krishnamachari, Bhaskar and Srivastava, Mani and Zhang, Mi},
title = {Artificial Intelligence of Things: A Survey},
year = {2024},
publisher = {Association for Computing Machinery},
address = {New York, NY, USA},
issn = {1550-4859},
url = {https://doi.org/10.1145/3690639},
doi = {10.1145/3690639},
note = {Just Accepted},
journal = {ACM Trans. Sen. Netw.},
month = aug
}

@article{Li2024Oct,
	author = {Li, Zechen and Deldari, Shohreh and Chen, Linyao and Xue, Hao and Salim, Flora D.},
	title = {{SensorLLM: Aligning Large Language Models with Motion Sensors for Human Activity Recognition}},
	journal = {arXiv},
	year = {2024},
	month = oct,
	eprint = {2410.10624},
	doi = {10.48550/arXiv.2410.10624}
}

@article{Imran2024Jun,
	author = {Imran, Sheikh Asif and Khan, Mohammad Nur Hossain and Biswas, Subrata and Islam, Bashima},
	title = {{LLaSA: Large Multimodal Agent for Human Activity Analysis Through Wearable Sensors}},
	journal = {arXiv},
	year = {2024},
	month = jun,
	eprint = {2406.14498},
	doi = {10.48550/arXiv.2406.14498}
}

@article{ji2024hargpt,
  title={HARGPT: Are LLMs Zero-Shot Human Activity Recognizers?},
  author={Ji, Sijie and Zheng, Xinzhe and Wu, Chenshu},
  journal={arXiv preprint arXiv:2403.02727},
  year={2024}
}

@article{liu2024visual,
  title={Visual instruction tuning},
  author={Liu, Haotian and Li, Chunyuan and Wu, Qingyang and Lee, Yong Jae},
  journal={Advances in neural information processing systems},
  volume={36},
  year={2024}
}

@article{hu2021lora,
  title={Lora: Low-rank adaptation of large language models},
  author={Hu, Edward J and Shen, Yelong and Wallis, Phillip and Allen-Zhu, Zeyuan and Li, Yuanzhi and Wang, Shean and Wang, Lu and Chen, Weizhu},
  journal={arXiv preprint arXiv:2106.09685},
  year={2021}
}

@inproceedings{gao2024unsupervised,
  title={Unsupervised Human Activity Recognition Via Large Language Models and Iterative Evolution},
  author={Gao, Jiayuan and Zhang, Yingwei and Chen, Yiqiang and Zhang, Tengxiang and Tang, Boshi and Wang, Xiaoyu},
  booktitle={ICASSP 2024-2024 IEEE International Conference on Acoustics, Speech and Signal Processing (ICASSP)},
  pages={91--95},
  year={2024},
  organization={IEEE}
}

@inproceedings{gardner2023llark,
  title={LLark: A Multimodal Instruction-Following Language Model for Music},
  author={Gardner, Joshua P and Durand, Simon and Stoller, Daniel and Bittner, Rachel M},
  booktitle={Forty-first International Conference on Machine Learning},
  year={2023}
}

@article{yuan2024self,
  title={Self-supervised learning for human activity recognition using 700,000 person-days of wearable data},
  author={Yuan, Hang and Chan, Shing and Creagh, Andrew P and Tong, Catherine and Acquah, Aidan and Clifton, David A and Doherty, Aiden},
  journal={NPJ digital medicine},
  volume={7},
  number={1},
  pages={91},
  year={2024},
  publisher={Nature Publishing Group UK London}
}

@article{Willetts2018May,
	author = {Willetts, Matthew and Hollowell, Sven and Aslett, Louis and Holmes, Chris and Doherty, Aiden},
	title = {{Statistical machine learning of sleep and physical activity phenotypes from sensor data in 96,220 UK Biobank participants}},
	journal = {Sci. Rep.},
	volume = {8},
	number = {7961},
	pages = {1--10},
	year = {2018},
	month = may,
	issn = {2045-2322},
	publisher = {Nature Publishing Group},
	doi = {10.1038/s41598-018-26174-1}
}

@misc{song2022mlbasedsecurelowpowercommunication,
      title={ML-based Secure Low-Power Communication in Adversarial Contexts}, 
      author={Guanqun Song and Ting Zhu},
      year={2022},
      eprint={2212.13689},
      archivePrefix={arXiv},
      primaryClass={cs.CR},
      url={https://arxiv.org/abs/2212.13689}, 
}

@misc{gopal2022securityprivacychallengesmicroservices,
      title={Security, Privacy and Challenges in Microservices Architecture and Cloud Computing- Survey}, 
      author={Hemanth Gopal and Guanqun Song and Ting Zhu},
      year={2022},
      eprint={2212.14422},
      archivePrefix={arXiv},
      primaryClass={cs.CR},
      url={https://arxiv.org/abs/2212.14422}, 
}

@misc{safavi2023efficientsemanticsegmentationedge,
      title={Efficient Semantic Segmentation on Edge Devices}, 
      author={Farshad Safavi and Irfan Ali and Venkatesh Dasari and Guanqun Song and Ting Zhu and Maryam Rahnemoonfar},
      year={2023},
      eprint={2212.13691},
      archivePrefix={arXiv},
      primaryClass={cs.CV},
      url={https://arxiv.org/abs/2212.13691}, 
}

@misc{khatri2022heterogeneouscomputingsystems,
      title={Heterogeneous Computing Systems}, 
      author={Dimple P. Khatri and Guanqun Song and Ting Zhu},
      year={2022},
      eprint={2212.14418},
      archivePrefix={arXiv},
      primaryClass={eess.SY},
      url={https://arxiv.org/abs/2212.14418}, 
}

@misc{ketha2025analysissecurityoslevelvirtualization,
      title={Analysis of Security in OS-Level Virtualization}, 
      author={Krishna Sai Ketha and Guanqun Song and Ting Zhu},
      year={2025},
      eprint={2501.01334},
      archivePrefix={arXiv},
      primaryClass={cs.CR},
      url={https://arxiv.org/abs/2501.01334}, 
}

@misc{shergill2024energyefficientlorawanleo,
      title={Energy Efficient LoRaWAN in LEO Satellites}, 
      author={Muskan Shergill and Zach Thompson and Guanqun Song and Ting Zhu},
      year={2024},
      eprint={2412.20660},
      archivePrefix={arXiv},
      primaryClass={cs.ET},
      url={https://arxiv.org/abs/2412.20660}, 
}

@misc{ali2023security5gnetworks,
      title={Security in 5G Networks -- How 5G networks help Mitigate Location Tracking Vulnerability}, 
      author={Abshir Ali and Guanqun Song and Ting Zhu},
      year={2023},
      eprint={2312.16200},
      archivePrefix={arXiv},
      primaryClass={cs.CR},
      url={https://arxiv.org/abs/2312.16200}, 
}

@misc{gould2024environmentaleconomicimpactio,
      title={Environmental and Economic Impact of I/O Device Obsolescence}, 
      author={Patrick Gould and Guanqun Song and Ting Zhu},
      year={2024},
      eprint={2412.20655},
      archivePrefix={arXiv},
      primaryClass={cs.CY},
      url={https://arxiv.org/abs/2412.20655}, 
}

@misc{gao2024optimizingglobalquantumcommunication,
      title={Optimizing Global Quantum Communication via Satellite Constellations}, 
      author={Yichen Gao and Guanqun Song and Ting Zhu},
      year={2024},
      eprint={2501.00280},
      archivePrefix={arXiv},
      primaryClass={quant-ph},
      url={https://arxiv.org/abs/2501.00280}, 
}

@misc{yuan2024heatsatellitesmeatgpus,
      title={Heat: Satellite's meat is GPU's poison}, 
      author={Zhehu Yuan and Jinyang Liu and Guanqun Song and Ting Zhu},
      year={2024},
      eprint={2501.14757},
      archivePrefix={arXiv},
      primaryClass={cs.DC},
      url={https://arxiv.org/abs/2501.14757}, 
}

@misc{kulshrestha2023innerworkingswindowssecurity,
      title={The Inner Workings of Windows Security}, 
      author={Ashvini A Kulshrestha and Guanqun Song and Ting Zhu},
      year={2023},
      eprint={2312.15150},
      archivePrefix={arXiv},
      primaryClass={cs.CR},
      url={https://arxiv.org/abs/2312.15150}, 
}

@misc{dixit2023dataclassificationmultiprocessing,
      title={Data Classification With Multiprocessing}, 
      author={Anuja Dixit and Shreya Byreddy and Guanqun Song and Ting Zhu},
      year={2023},
      eprint={2312.15152},
      archivePrefix={arXiv},
      primaryClass={cs.LG},
      url={https://arxiv.org/abs/2312.15152}, 
}

@misc{li2022minisculesurveyblockchainscalability,
      title={A Miniscule Survey on Blockchain Scalability}, 
      author={Angela Li and Guanqun Song and Ting Zhu},
      year={2022},
      eprint={2212.13353},
      archivePrefix={arXiv},
      primaryClass={cs.CR},
      url={https://arxiv.org/abs/2212.13353}, 
}

@misc{yu2024achievingcarbonneutralityio,
      title={Achieving Carbon Neutrality for I/O Devices}, 
      author={Botao Yu and Guanqun Song and Ting Zhu},
      year={2024},
      eprint={2501.14774},
      archivePrefix={arXiv},
      primaryClass={cs.CY},
      url={https://arxiv.org/abs/2501.14774}, 
}

@misc{cheng2024technologicalprogressobsolescenceanalyzing,
      title={Technological Progress and Obsolescence: Analyzing the Environmental Economic Impacts of MacBook Pro I/O Devices}, 
      author={Yun-Chieh Cheng and Yu-Tong Shen and Guanqun Song and Ting Zhu},
      year={2024},
      eprint={2501.14758},
      archivePrefix={arXiv},
      primaryClass={cs.CY},
      url={https://arxiv.org/abs/2501.14758}, 
}

@misc{sun2023designimplementationconsiderationsvirtual,
      title={Design and Implementation Considerations for a Virtual File System Using an Inode Data Structure}, 
      author={Qin Sun and Grace McKenzie and Guanqun Song and Ting Zhu},
      year={2023},
      eprint={2312.15153},
      archivePrefix={arXiv},
      primaryClass={cs.OS},
      url={https://arxiv.org/abs/2312.15153}, 
}

@misc{qiu2023mapreducemultiprocessinglargedata,
      title={Map-Reduce for Multiprocessing Large Data and Multi-threading for Data Scraping}, 
      author={Zefeng Qiu and Prashanth Umapathy and Qingquan Zhang and Guanqun Song and Ting Zhu},
      year={2023},
      eprint={2312.15158},
      archivePrefix={arXiv},
      primaryClass={math.NA},
      url={https://arxiv.org/abs/2312.15158}, 
}

@article{YAO2020100087,
title = {Paris: Passive and continuous fetal heart monitoring system},
journal = {Smart Health},
volume = {17},
pages = {100087},
year = {2020},
issn = {2352-6483},
doi = {https://doi.org/10.1016/j.smhl.2019.100087},
url = {https://www.sciencedirect.com/science/article/pii/S2352648319300510},
author = {Yao Yao and Zeyu Ning and Qingquan Zhang and Ting Zhu}
}

@misc{wei2025,
      title={Self-Consuming Generative Models with Adversarially Curated Data}, 
      author={Xiukun Wei and Xueru Zhang},
      year={2025},
      eprint={2505.09768},
      archivePrefix={arXiv},
      primaryClass={cs.LG},
      url={https://arxiv.org/abs/2505.09768}, 
}

@INPROCEEDINGS{9894326,
  author={Wei, Xiaohui and Wei, Xiukun and Wang, Xingwang and Wang, Yundi and Niu, Yan},
  booktitle={2022 IEEE International Performance, Computing, and Communications Conference (IPCCC)}, 
  title={HRCache: Edge-End Collaboration for Mobile Deep Vision Based on H.264 and Approximated Reuse}, 
  year={2022},
  volume={},
  number={},
  pages={380-388},
  doi={10.1109/IPCCC55026.2022.9894326}}

@article{MILLER2022100245,
title = {Radar-based monitoring system for medication tampering using data augmentation and multivariate time series classification},
journal = {Smart Health},
volume = {23},
pages = {100245},
year = {2022},
issn = {2352-6483},
doi = {https://doi.org/10.1016/j.smhl.2021.100245},
url = {https://www.sciencedirect.com/science/article/pii/S235264832100060X},
author = {Elishiah Miller and Zane MacFarlane and Seth Martin and Nilanjan Banerjee and Ting Zhu}
}

@inproceedings{wire1,
author = {Wang, Wei and Liu, Xin and Chi, Zicheng and Ray, Stuart and Zhu, Ting},
title = {Key Establishment for Secure Asymmetric Cross-Technology Communication},
year = {2024},
isbn = {9798400704826},
publisher = {Association for Computing Machinery},
address = {New York, NY, USA},
url = {https://doi-org.proxy.lib.ohio-state.edu/10.1145/3634737.3637670},
doi = {10.1145/3634737.3637670},
booktitle = {Proceedings of the 19th ACM Asia Conference on Computer and Communications Security},
pages = {412–422},
numpages = {11},
location = {Singapore, Singapore},
series = {ASIA CCS '24}
}

@inproceedings {wire2,
author = {Xin Liu and Wei Wang and Guanqun Song and Ting Zhu},
title = {{LightThief}: Your Optical Communication Information is Stolen behind the Wall},
booktitle = {32nd USENIX Security Symposium (USENIX Security 23)},
year = {2023},
isbn = {978-1-939133-37-3},
address = {Anaheim, CA},
pages = {5325--5339},
url = {https://www.usenix.org/conference/usenixsecurity23/presentation/liu-xin},
publisher = {USENIX Association},
month = aug
}

@misc{wire3,
      title={ML-based Secure Low-Power Communication in Adversarial Contexts}, 
      author={Guanqun Song and Ting Zhu},
      year={2022},
      eprint={2212.13689},
      archivePrefix={arXiv},
      primaryClass={cs.CR},
      url={https://arxiv.org/abs/2212.13689}, 
}

@article{GAO201718,
title = {A smart medical system for dynamic closed-loop blood glucose-insulin control},
journal = {Smart Health},
volume = {1-2},
pages = {18-33},
year = {2017},
note = {Connected Health: Applications, Systems and Engineering Technologies (CHASE 2016)},
issn = {2352-6483},
doi = {https://doi.org/10.1016/j.smhl.2017.04.001},
url = {https://www.sciencedirect.com/science/article/pii/S2352648317300119},
author = {Jialin Gao and Ping Yi and Zicheng Chi and Ting Zhu}
}

@article{MILLER2020100089,
title = {RadSense: Enabling one hand and no hands interaction for sterile manipulation of medical images using Doppler radar},
journal = {Smart Health},
volume = {15},
pages = {100089},
year = {2020},
issn = {2352-6483},
doi = {https://doi.org/10.1016/j.smhl.2019.100089},
url = {https://www.sciencedirect.com/science/article/pii/S2352648319300534},
author = {Elishiah Miller and Zheng Li and Helena Mentis and Adrian Park and Ting Zhu and Nilanjan Banerjee}
}

@inproceedings{10.1145/3460120.3484766,
author = {Wang, Wei and Yao, Yao and Liu, Xin and Li, Xiang and Hao, Pei and Zhu, Ting},
title = {I Can See the Light: Attacks on Autonomous Vehicles Using Invisible Lights},
year = {2021},
isbn = {9781450384544},
publisher = {Association for Computing Machinery},
address = {New York, NY, USA},
url = {https://doi-org.proxy.lib.ohio-state.edu/10.1145/3460120.3484766},
doi = {10.1145/3460120.3484766},
booktitle = {Proceedings of the 2021 ACM SIGSAC Conference on Computer and Communications Security},
pages = {1930–1944},
numpages = {15},
location = {Virtual Event, Republic of Korea},
series = {CCS '21}
}

@ARTICLE{9340574,
  author={Chi, Zicheng and Li, Yan and Sun, Hongyu and Huang, Zhichuan and Zhu, Ting},
  journal={IEEE/ACM Transactions on Networking}, 
  title={Simultaneous Bi-Directional Communications and Data Forwarding Using a Single ZigBee Data Stream}, 
  year={2021},
  volume={29},
  number={2},
  pages={821-833},
  doi={10.1109/TNET.2021.3054339}}

@inproceedings{10.1145/3387514.3405861,
author = {Chi, Zicheng and Liu, Xin and Wang, Wei and Yao, Yao and Zhu, Ting},
title = {Leveraging Ambient LTE Traffic for Ubiquitous Passive Communication},
year = {2020},
isbn = {9781450379557},
publisher = {Association for Computing Machinery},
address = {New York, NY, USA},
url = {https://doi-org.proxy.lib.ohio-state.edu/10.1145/3387514.3405861},
doi = {10.1145/3387514.3405861},
booktitle = {Proceedings of the Annual Conference of the ACM Special Interest Group on Data Communication on the Applications, Technologies, Architectures, and Protocols for Computer Communication},
pages = {172–185},
numpages = {14},
location = {Virtual Event, USA},
series = {SIGCOMM '20}
}

@INPROCEEDINGS{9120764,
  author={Tao, Yinrong and Xiao, Sheng and Hao, Bin and Zhang, Qingquan and Zhu, Ting and Chen, Zhuo},
  booktitle={2020 IEEE Wireless Communications and Networking Conference (WCNC)}, 
  title={WiRE: Security Bootstrapping for Wireless Device-to-Device Communication}, 
  year={2020},
  volume={},
  number={},
  pages={1-7},
  doi={10.1109/WCNC45663.2020.9120764}}

@inproceedings{10.1145/3356250.3360046,
author = {Chi, Zicheng and Li, Yan and Liu, Xin and Yao, Yao and Zhang, Yanchao and Zhu, Ting},
title = {Parallel inclusive communication for connecting heterogeneous IoT devices at the edge},
year = {2019},
isbn = {9781450369503},
publisher = {Association for Computing Machinery},
address = {New York, NY, USA},
url = {https://doi-org.proxy.lib.ohio-state.edu/10.1145/3356250.3360046},
doi = {10.1145/3356250.3360046},
booktitle = {Proceedings of the 17th Conference on Embedded Networked Sensor Systems},
pages = {205–218},
numpages = {14},
location = {New York, New York},
series = {SenSys '19}
}

@ARTICLE{8694952,
  author={Chi, Zicheng and Li, Yan and Sun, Hongyu and Yao, Yao and Zhu, Ting},
  journal={IEEE/ACM Transactions on Networking}, 
  title={Concurrent Cross-Technology Communication Among Heterogeneous IoT Devices}, 
  year={2019},
  volume={27},
  number={3},
  pages={932-947},
  doi={10.1109/TNET.2019.2908754}}

@inproceedings{10.1145/3274783.3274846,
author = {Li, Yan and Chi, Zicheng and Liu, Xin and Zhu, Ting},
title = {Passive-ZigBee: Enabling ZigBee Communication in IoT Networks with 1000X+ Less Power Consumption},
year = {2018},
isbn = {9781450359528},
publisher = {Association for Computing Machinery},
address = {New York, NY, USA},
url = {https://doi-org.proxy.lib.ohio-state.edu/10.1145/3274783.3274846},
doi = {10.1145/3274783.3274846},
booktitle = {Proceedings of the 16th ACM Conference on Embedded Networked Sensor Systems},
pages = {159–171},
numpages = {13},
location = {Shenzhen, China},
series = {SenSys '18}
}

@inproceedings{10.1145/3210240.3210346,
author = {Li, Yan and Chi, Zicheng and Liu, Xin and Zhu, Ting},
title = {Chiron: Concurrent High Throughput Communication for IoT Devices},
year = {2018},
isbn = {9781450357203},
publisher = {Association for Computing Machinery},
address = {New York, NY, USA},
url = {https://doi-org.proxy.lib.ohio-state.edu/10.1145/3210240.3210346},
doi = {10.1145/3210240.3210346},
booktitle = {Proceedings of the 16th Annual International Conference on Mobile Systems, Applications, and Services},
pages = {204–216},
numpages = {13},
location = {Munich, Germany},
series = {MobiSys '18}
}

@INPROCEEDINGS{8486349,
  author={Wang, Wei and Xie, Tiantian and Liu, Xin and Zhu, Ting},
  booktitle={IEEE INFOCOM 2018 - IEEE Conference on Computer Communications}, 
  title={ECT: Exploiting Cross-Technology Concurrent Transmission for Reducing Packet Delivery Delay in IoT Networks}, 
  year={2018},
  volume={},
  number={},
  pages={369-377},
  doi={10.1109/INFOCOM.2018.8486349}}

@ARTICLE{9444204,
  author={Han, Dianqi and Li, Ang and Zhang, Lili and Zhang, Yan and Li, Jiawei and Li, Tao and Zhu, Ting and Zhang, Yanchao},
  journal={IEEE/ACM Transactions on Networking}, 
  title={Deep Learning-Guided Jamming for Cross-Technology Wireless Networks: Attack and Defense}, 
  year={2021},
  volume={29},
  number={5},
  pages={1922-1932},
  doi={10.1109/TNET.2021.3082839}}

@misc{ning2021benchmarkingmachinelearningfast,
      title={Benchmarking Machine Learning: How Fast Can Your Algorithms Go?}, 
      author={Zeyu Ning and Hugues Nelson Iradukunda and Qingquan Zhang and Ting Zhu},
      year={2021},
      eprint={2101.03219},
      archivePrefix={arXiv},
      primaryClass={cs.LG},
      url={https://arxiv.org/abs/2101.03219}, 
}

@INPROCEEDINGS{8556807,
  author={Meng, Yishuang and Yi, Ping and Guo, Xuejun and Gu, Wen and Liu, Xin and Wang, Wei and Zhu, Ting},
  booktitle={2018 Third International Conference on Security of Smart Cities, Industrial Control System and Communications (SSIC)}, 
  title={Detection for Pulmonary Nodules using RGB Channel Superposition Method in Deep Learning Framework}, 
  year={2018},
  volume={},
  number={},
  pages={1-8},
  doi={10.1109/SSIC.2018.8556807}}

@ARTICLE{10189210,
  author={Liu, Xin and Chi, Zicheng and Wang, Wei and Yao, Yao and Hao, Pei and Zhu, Ting},
  journal={IEEE/ACM Transactions on Networking}, 
  title={High-Granularity Modulation for OFDM Backscatter}, 
  year={2024},
  volume={32},
  number={1},
  pages={338-351},
  doi={10.1109/TNET.2023.3286880}}

@ARTICLE{10125074,
  author={Yao, Yao and Li, Yan and Zhu, Ting},
  journal={IEEE Transactions on Mobile Computing}, 
  title={Interference-Negligible Privacy-Preserved Shield for RF Sensing}, 
  year={2024},
  volume={23},
  number={5},
  pages={3576-3588},
  doi={10.1109/TMC.2023.3276930}}

@inproceedings {285483,
author = {Ang Li and Jiawei Li and Dianqi Han and Yan Zhang and Tao Li and Ting Zhu and Yanchao Zhang},
title = {{PhyAuth}: {Physical-Layer} Message Authentication for {ZigBee} Networks},
booktitle = {32nd USENIX Security Symposium (USENIX Security 23)},
year = {2023},
isbn = {978-1-939133-37-3},
address = {Anaheim, CA},
pages = {1--18},
url = {https://www.usenix.org/conference/usenixsecurity23/presentation/li-ang},
publisher = {USENIX Association},
month = aug
}

@inproceedings{10.1145/3395351.3399367,
author = {Chi, Zicheng and Li, Yan and Liu, Xin and Wang, Wei and Yao, Yao and Zhu, Ting and Zhang, Yanchao},
title = {Countering cross-technology jamming attack},
year = {2020},
isbn = {9781450380065},
publisher = {Association for Computing Machinery},
address = {New York, NY, USA},
url = {https://doi-org.proxy.lib.ohio-state.edu/10.1145/3395351.3399367},
doi = {10.1145/3395351.3399367},
booktitle = {Proceedings of the 13th ACM Conference on Security and Privacy in Wireless and Mobile Networks},
pages = {99–110},
numpages = {12},
location = {Linz, Austria},
series = {WiSec '20}
}

@inproceedings{10.1145/3769102.3770620,
author = {Song, Guanqun and Li, Yan and Zhu, Ting},
title = {A Metal Sensing and Biometric-based Tracking System},
year = {2025},
isbn = {9798400722387},
publisher = {Association for Computing Machinery},
address = {New York, NY, USA},
url = {https://doi-org.proxy.lib.ohio-state.edu/10.1145/3769102.3770620},
doi = {10.1145/3769102.3770620},
booktitle = {Proceedings of the Tenth ACM/IEEE Symposium on Edge Computing},
articleno = {23},
numpages = {17},
location = {the Hilton Arlington National Landing, Arlington, VA, USA},
series = {SEC '25}
}

\end{document}